\documentclass{article}
\usepackage[utf8]{inputenc}

\usepackage{authblk}
\usepackage[nonatbib, preprint]{neurips_2022} 
\usepackage[numbers]{natbib}
\usepackage{subcaption}
\relax
\usepackage{graphicx}
\usepackage{multirow}
\usepackage{graphicx}
\usepackage{subcaption}
\usepackage{booktabs}
\usepackage{amsmath}
\usepackage{tikz}
\usepackage{wrapfig}
\usetikzlibrary{arrows}
\usepackage{float}

\usepackage{amsthm}
\numberwithin{equation}{section} 

\newenvironment{itemizeReduced}{
\begin{list}{\labelitemi}{\leftmargin=1.5em}
\setlength{\itemsep}{0.6pt}
\setlength{\parskip}{0pt}
\setlength{\parsep}{0pt}}{\end{list}
}
\renewcommand\labelitemi{$\vcenter{\hbox{\tiny$\bullet$}}$}
\usepackage{tabularx}

\usepackage[verbose=true,letterpaper]{geometry}
\AtBeginDocument{
  \newgeometry{
    textheight=9in,
    textwidth=5.5in,
    top=1in,
    headheight=12pt,
    headsep=25pt,
    footskip=30pt
  }
}
\usepackage{parskip}
\usepackage{hyperref}
\usepackage{fancyhdr}
\usepackage{amsfonts,amssymb}
\usepackage[titlenumbered,ruled]{algorithm2e}
\fancyheadoffset{0pt}
\title{Neural Payoff Machines: Predicting Fair and Stable Payoff Allocations Among Team Members}
\author[1]{Daphne Cornelisse}
\author[1]{Thomas Rood}
\author[2]{Mateusz Malinowski}
\author[2]{Yoram Bachrach}
\author[1,3]{Tal Kachman}

\affil[1]{Department of Artificial Intelligence \\ Radboud University, Nijmegen, the Netherlands

}
\affil[3]{Donders Institute for Brain, Cognition and Behaviour \\ Radboud University, Nijmegen, the Netherlands}
\affil[2]{DeepMind}
\date{\today}

\begin{document}
\maketitle

\begin{abstract}
In many multi-agent settings, participants can form teams to achieve collective outcomes that may far surpass their individual capabilities. Measuring the relative contributions of agents and allocating them shares of the reward that promote long-lasting cooperation are difficult tasks. Cooperative game theory offers solution concepts identifying distribution schemes, such as the Shapley value, that fairly reflect the contribution of individuals to the performance of the team or the Core, which reduces the incentive of agents to abandon their team. Applications of such methods include identifying influential features and sharing the costs of joint ventures or team formation. Unfortunately, using these solutions requires tackling a computational barrier as they are hard to compute, even in restricted settings. In this work, we show how cooperative game-theoretic solutions can be distilled into a learned model by training neural networks to propose fair and stable payoff allocations. We show that our approach creates models that can generalize to games far from the training distribution and can predict solutions for more players than observed during training. An important application of our framework is Explainable AI: our approach can be used to speed-up Shapley value computations on many instances.
\end{abstract}

\section{Introduction}

The ability of individuals to form teams and collaborate is crucial to their performance in many environments. The success of humans as a species hinges on our capability to cooperate at scale~\cite{henrich2015secret}. Similarly, cooperation between learning agents is necessary to achieve high performance in many environments~\cite{lowe2017multi,baker2019emergent,vinyals2019grandmaster,jaderberg2019human} and is a fundamental problem in artificial intelligence~\cite{stone2000multiagent,dafoe2021cooperative}. 
Individual agents are often not incentivized by the joint reward achieved by a team but rather by their share of the spoils. Hence, teams are only likely to be formed when the overall gains obtained by the team are appropriately distributed between its members.  However, understanding how collective outcomes arise from subsets of locally interacting parts, or measuring the impact of individuals on the team's performance, remain open problems.

Direct applications exist in multiple domains. One example is identifying the most influential features that drive a model to make a certain prediction~\cite{datta2016algorithmic,lundberg2017unified,bachrach2012crowd,sundararajan2017axiomatic,lorraine2021lyapunov,lorraine2021using}; one of the cornerstones of explainable AI~\cite{arrieta2020explainable,metz2021gradients}. Another example is sharing the costs of data acquisition or a joint venture in a fair way between participants~\cite{balkanski2017statistical,agarwal2019marketplace}, or sharing gains between cooperating agents~\cite{gately1974sharing,sim2020collaborative}. In many legislative bodies individual participants have different weights, and passing a decision requires support from a set of participants holding the majority of the weight; different states in the US electoral college have different numbers of electors, and different countries in the EU Council of Ministers vary in their voting weight. Here, would like to quantify the true political power held by each participant, or allocate a common budget between them~\cite{mann1962values,bilbao2002voting}. 


Cooperative game theory can provide strong theoretical foundations underpinning such applications. The field provides solution concepts that measure the relative impact of individuals on team performance, or the individual rewards agents are entitled to. \textit{Power indices} such as the Shapley value~\cite{shapley1953value} or Banzhaf index~\cite{banzhaf1964weighted} attempt to divide the joint reward in a way that is \textit{fair}, and have recently been used to compute feature importance \cite{lundberg2017unified}. In contrast, other solutions such as the Core~\cite{gillies1953some} attempt to offer a {\it stable} allocation of payoffs, where individual agents are incentivised to continue working with their team, rather than attempting to break away from their team in favor of working with other agents. Despite their theoretical appeal, these solution concepts are difficult to apply in practice due to computational constraints. Computing them is typically a hard problem, even in restricted environments~\cite{elkind2007computational,chalkiadakis2011computational,deng1994complexity}. 

\begin{figure}[ht]
    \centering
    \includegraphics[width=0.75\textwidth]{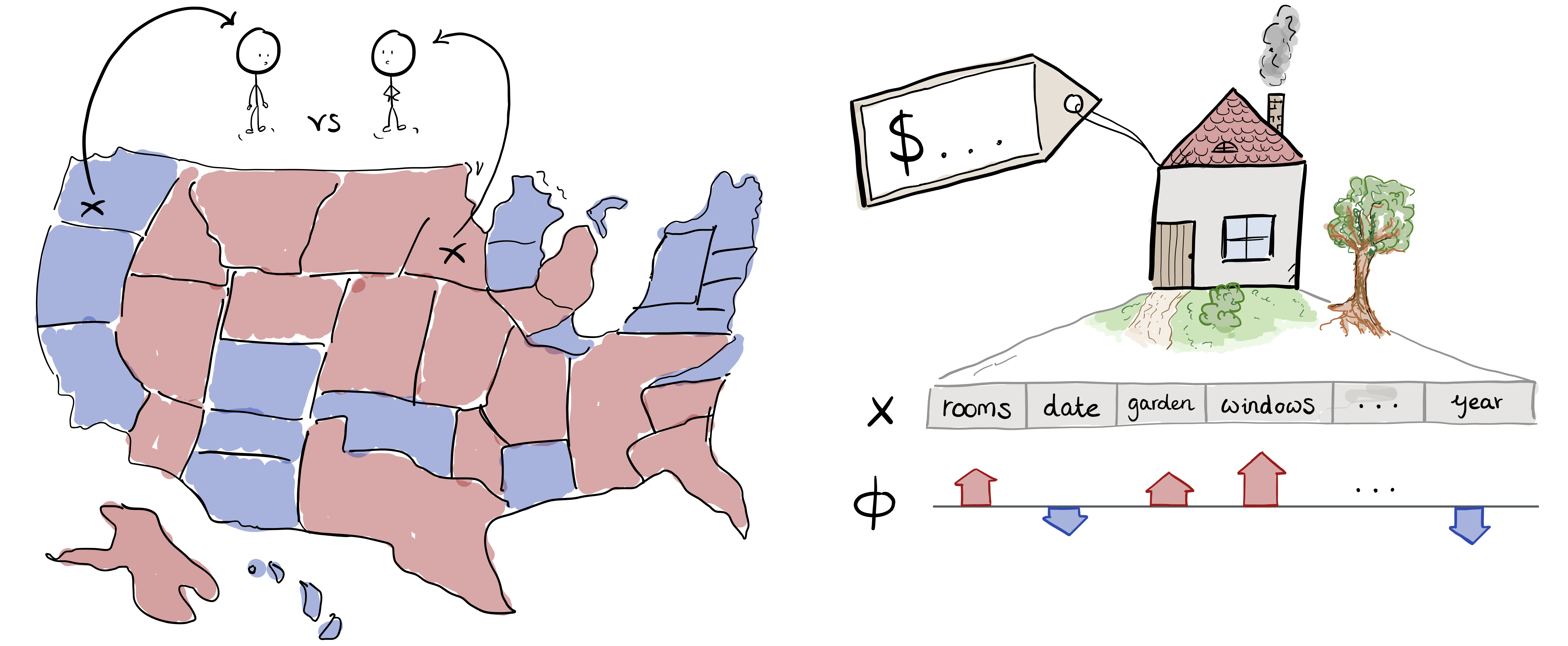}
    \caption{\textbf{Evaluation domains for our approach.} \textit{Left}: Weighted voting games (WVGs) model decision making bodies such as the US Electoral College ~\cite{mann1962values,bilbao2002voting}. Our approach predicts solution concepts which reflect the true political influence of players or stable payoff allocations. \textit{Right}: Applying the Shapley value in Feature Importance Games enables quantifying the relative impact of features on the decisions of a model. In this example, a model predicts the price of a house based on several features. Our approach predicts the Shapley values $\boldsymbol{\phi}$ of features, which are a commonly used metric for their impact. This speeds-up Explainable AI analysis, as each new instance can be analyzed quickly without a full computation of Shapley values.}
    \label{fig:examples}
\end{figure}

{\bf Our contribution:} We construct models that predict fair or stable payoff allocations among team members, combining solution concepts from cooperative game theory with the predictive power of neural networks. These neural ``payoff machines'' take in a representation of the performance or reward achievable by different subsets of agents, and output a suggested payoff vector allocating the total reward between the agents. By training the neural networks based on different cooperative solution concepts, the model can be tuned to aim for a fair distribution of the payoffs (the Shapley value or Banzhaf index) or to minimize the incentives of agents to abandon the team (the Least-Core~\cite{gillies1953some,maschler1979geometric,deng1994complexity}).  
Figure~\ref{fig:examples} depicts the two well-studied classes of games on which we evaluate our approach: weighted voting games~\cite{mann1962values,bilbao2002voting,chalkiadakis2011computational} and feature importance games in explainable AI~\cite{datta2016algorithmic,lundberg2017unified}.

{\bf Weighted voting games (WVGs)} are arguably the most well-studied class of cooperative games. Each agent is endowed with a weight and a team achieves its goal if the sum of the weights of the team members exceeds a certain threshold (quota). We train the neural networks by generating large sets of such games and computing their respective game theoretic solutions. Our empirical evaluation shows that the predictions for the various solutions (the Shapley value, Banzhaf index and Least-Core) accurately reflect the true game theoretic solutions on previously unobserved games. Furthermore, the resulting model can generalize even to games that are very far from the training distribution or with more players than the games in the training set. 

{\bf Feature importance games} are a model for quantifying the relative influence of features on the outcome of a machine learning model~\cite{datta2016algorithmic,lundberg2017unified}. Solving these games for the Shapley value (or other game theoretic measure) provides a way to reverse-engineer the key factors that drove a model to reach a specific  decision. This approach is model-agnostic, thus can be applied to make any ``black-box'' model more interpretable~\cite{arrieta2020explainable}. One drawback of this approach is the computational complexity of calculating the Shapley value, making such analysis slow even when using approximation algorithms. Our approach provides a way to significantly speedup Explainable AI analyses, particularly for datasets with a large number of instances.

\section{Preliminaries}
We provide a brief overview of cooperative game theory (examined in details in various books~\cite{osborne1994course,chalkiadakis2011computational}) and discuss how solution concepts in cooperative game theory have been applied in Explainable AI~\cite{datta2016algorithmic,lundberg2017unified,lundberg2020local,das2020opportunities,yan2021if}.

\subsection{Cooperative Game Theory} \label{cooperativegametheory} 
A (transferable utility) {\bf cooperative game} consists of a set $N = \{1, 2, \dots, n \}$ of agents, or players, and a characteristic function $v:2^{n} \to \mathbb{R}$ which maps each team of players, or \textit{coalition} $C \subseteq N$, to a real number. This number indicates the joint reward the players obtained as a team. Games $v$ where $v:2^{n} \to \{0, 1\}$ (binary range) are {\it simple} games.

\textbf{Weighted voting games} (WVGs) are a restricted class of simple cooperative games \cite{chalkiadakis2011computational}, where each agent $i$ has a weight $w_i$ and a team of agents $C \subseteq N$ wins if the sum of the weights of its participants $\sum_{i=1}^n w_i$ exceeds a quota $q$. Formally, a WVG is defined as the triple $(\mathbf{w}, q, v)$ with weights $\mathbf{w} = (w_1, w_2, \dots, w_n) \in \mathbb{R}_{\geq 0}^{n}$ and quota (threshold) $q \in \mathbb{R}_{\geq 0}$ where for any $C \subseteq N$ we have $v(C)=1$ if $\sum_{i=1}^n w_i \geq q$ and otherwise $v(C)=0$. If $v(C) = 0$ we say $C$ is a losing coalition, and if $v(C)=1$ we say it is a winning coalitions. WVGs have been thoroughly investigated as a model of voting bodies, such the US Electoral College or the EU Council of Ministers~\cite{mann1962values,bilbao2002voting}. 

The characteristic function defines the joint value of a coalition, but it does not specify how the value should be distributed among the agents. {\it Solution concepts} attempt to determine an allocation $\mathbf{p} = (p_1, \ldots, p_n)$ of the utility $v(N)$ achieved by the grand coalition of all the agents; an allocation $\mathbf{p}$ is called an {\it imputation} if for any player $i$ we have $p_i \geq 0$ and ${\sum_{i =1}^n p_i = v(N)}$. 
\footnote{In a WVG, the value of a coalition is bounded by 1, so if the grand coalition $N$ indeed has a value of $v(N)=1$ then a solution would be a payoff vector $\mathbf{p} = (p_1, \ldots, p_n)$ where $\sum_{i = 1}^n p_i = 1$.}
This allocation is meant to achieve some desiderata, such as fairly reflecting the contributions of individual agents, or achieving stability in the sense that no subset of agents is incentivized to abandon the team and form a new team. We describe three prominent solution concepts that we use in our analysis.

\paragraph{The Core.} Rational players may abandon the grand coalition of all the agents if they can increase their individual utility by doing so. The Core is defined as the set of all payoff vectors where no subset of agents can generate more utility, as measured by the characteristic function, than the total payoff they are currently awarded by the payoff vector. As such, the Core is viewed as the set of \textit{stable} payoff allocations. Formally, the core \cite{gillies1953some} is defined as the set of all imputations $\mathbf{p}$ such that $\sum_{i=1}^n p_i = v(N)$ and that $\sum_{i \in C} p_i \geq v(C)$ for any coalition $C \subseteq N$.

\paragraph{The $\varepsilon$-Core and Least-Core}~\cite{maschler1979geometric,deng1994complexity}. Some games have empty cores, meaning that no payoff allocations achieves full stability (i.e. for any imputation $\mathbf{p}$ there exists at least one coalition $C$ such that $v(C) > \sum_{i \in C} p_i$). In such cases, researchers have proposed minimizing the instability. A relaxation of the core is the \textbf{$\boldsymbol{\varepsilon}$-core}, consisting of imputations $\mathbf{p}$ where for any value coalition $C$ we have $p(C) \geq v(C) - \varepsilon$. Given an imputation $\mathbf{p}$ the difference $v(C) - \sum_{i \in C} p_i$ is called the {\it excess} of the coalition, and represents the total improvement in utility the members of $C$ can achieve by abandoning the grand coalition and working on their own. For an imputation in the $\varepsilon$-core, no agent subset $C$ can achieve an addition of $\varepsilon$ in utility over the current total payoff offered to the team (i.e. no coalition has an excess of more than $\varepsilon$). The minimal $\varepsilon$ for which the $\varepsilon$-core is non-empty is called the {\it Least-Core Value} (LCV). The Least-Core minimizes the incentive of any agent subset to abandon the grand coalition, and the LCV thus represents the degree of instability (excess) under the imputation that best minimizes this instability. We find the set of payoffs associated with the LVC through linear programming (full details in Appendix~\ref{sec:algorithm_leastcore}). 

We now discuss two {\it power indices}, payoff distributions reflecting the true influence a player has on the performance of the team, that fairly allocate the total gains of the teams among the agents in it.

\paragraph{The Shapley Value}~\cite{shapley1953value} measures the average marginal contribution of each player across all permutations of the players. The Shapley value is the unique solution concept that fulfills several natural fairness axioms~\cite{dubey1975uniqueness}, and has thus found many applications from estimating feature importance~\cite{datta2016algorithmic,lundberg2017unified} to pruning neural networks~\cite{wang2021efficient,frankle2018lottery,ghorbani2020neuron}. Formally, we denote a permutation of the players by $\pi$ , where $\pi$ is a bijective mapping of $N$ to itself, and the set of all such permutations by $\Pi$. By $C_{\pi}(i)$ we denote all players appearing before $i$ in the permutation $\pi$. The Shapley value $\phi_i(v)$ of player $i$ is defined as:
\begin{align}
    \phi_i(v) = \frac{1}{N!} \sum_{\pi \in \Pi} \left[ v(C_{\pi}(i) \cup \{ i \}) - v(C_{\pi}(i)) \right]
\end{align}
Intuitively, one can consider starting with an empty coalition and adding the players' weights in the order of the permutation; the first player whose addition meets or exceeds the quota is considered the pivotal player in the permutation. The Shapley value then measures the proportion of permutations where a player is the pivotal player. 

\textbf{The Banzhaf index}~\cite{banzhaf1964weighted} is another method for distributing payoffs according to a players' ability to change the outcome of the game, but it reflects slightly different fairness axioms~\cite{straffin1988}. The Banzhaf index $\beta_i$ of a player $i$ is defined as the marginal contribution of a player across all {\it subsets} not containing that player:
\begin{align}
\beta_i(G) = \frac{1}{2^{N-1}} \sum_{C \subseteq N \setminus \{ i \}} \left[ v(C \, \cup \, \{ i \}) - v(C) \right]
\end{align}
In practice, we first compute the set of winning coalitions $C^{\text{win}} \subseteq C$ and count, for each player, the number of times it is critical or \textit{pivotal}, that is, $v(\{C \} \backslash {i}) = 0$. 

\subsection{Speeding Up Explainable AI In Large Datasets}
\label{l_sect_prelim_xai_shap}

An important goal of our work is to speedup Shapley/Core computations of many data instances. In machine learning, we train a model $f_{\boldsymbol{\theta}}$ to learn a mapping from some set of input features to an outcome. This could we can train a model to predict the price of a house based on a number of features, such as the number of rooms, the year it was built, and so on (Figure \ref{fig:examples}). In such settings, it is desirable, but challenging, to explain the model outputs in terms of the input features. Explainable AI addresses this issue, and recent years showed several applications of game theoretic metrics for measuring feature importance in the machine learning community \cite{sundararajan2017axiomatic, arrieta2020explainable}.

The fastest method to approximate Shapley values (also used in the SHAP package) is a Monte-Carlo approach \cite{lundberg2017unified}. A number of other methods exist whose runtime and accuracy depend on the number of samples used, usually on the order of several thousands \cite{leech2002computation, datta2016algorithmic, maleki2013bounding,bachrach2010approximating}. In a model setting, the characteristic function takes the value of the trained model output for a given instance: $f_{\boldsymbol{\theta^*}}(\mathbf{x})$. The Shapley value of a feature $i$ in a data instance $x$, $\phi_{x, i}$ is the effect that feature has on the model outcome. Sampling based-methods compute the contributions with respect to a base value, which is the average model output across all instances: $f_{\boldsymbol{\theta}^*}( x ) = \mathbb{E}[ f_{\boldsymbol{\theta}^*}( x )  ] + \sum_{i = 1}^n \phi_{x, i}$. Sampling based methods are not ideal for large datasets because they require a large number of re-evaluation samples per computation. We show how our approach can be employed to speedup Shapley or Core computations of many instances by training models to learn representations of feature attribution schemes. 

\section{Methods}
\label{l_sect_methods}
Our approach uses machine learning to create game-theoretic estimators. We generate synthetic datasets of games to train our models, spending compute up-front to allow for instant solutions afterwards. Our first domain concerns weighted voting games (WVGs) as they provide a generic framework for studying cooperation in multi-agent settings. 

Afterwards, we apply our approach to the Shapley-based feature importance setting. The main idea is that we can speed-up the computations of the relative impact of features on the predictions of a machine learning model. We do this by training models to approximate Shapley values of features, and examine their performance on previously unobserved instances.


\subsection{Weighted Voting Games}
\label{modelanddata}

We consider two types of models: \textit{fixed-size} models that always predict a fixed number of outputs for games with a fixed number of players, and \textit{variable-size} ones that predict a variable number of outcomes and can handle variable numbers of players. We now describe our data generation process, models, training procedure, and evaluation metrics used in weighted voting games.

\subsubsection{Data and Models: Fixed-Size} 
For each $n$ player game, we generate $G$ independent and identically distributed games. The training dataset is given by ${\mathcal{D}^n_{\text{fixed}} = \left\{ \mathbf{X} \in \mathbb{R}^{G \times n}, \mathbf{P} \in \mathbb{R}^{G \times K} \right\}}$ where $K$ denotes the number of outputs of interest. The features are obtained in two steps. First, we sample a weight vector $\mathbf{w} \sim \text{Beta}(\alpha=1, \beta=1)$ on the interval from 1 to $2n$, and a quota ${q \sim \mathcal{N}(\mu = \frac{1}{4}(2n + 1)n, \sigma^2 = 2n)}$. To create games where players are dependent on each other to achieve the task at hand, the quota distribution is parameterized such that the average drawn quota is half of the sum of the players' weights. We then normalize the weights by the quota to get the feature vector $\mathbf{x}=\frac{1}{q}\cdot \mathbf{w}$. In other words, we have $n$ features which are the weights normalized as the respective proportion of the quota. Thus, a value $x_i > 1$ indicates that player $i$ is a winning coalition by itself, and needs other players to meet the quota otherwise (Figure~\ref{fig:data_gen}).

We train models to predict the three solutions: the Least-Core, the Shapley values, and the Banzhaf indices. For the Shapey values and the Banzhaf indices, the model predicts the payoff allocation $(p_1, \ldots, p_n)$ so $K \equiv n$. For the Least-Core solution, we train the model to not only predict the payoff allocation $p_1, \ldots, p_n$, but also to predict the Least Core Value $\varepsilon_{min}$ (so in this case the model has $K = n+1$ outputs). For each $n$ player game, we produce a model $f_{\boldsymbol{\theta}}: \mathbb{R}^{n} \to \mathbb{R}^{K}$, where $\boldsymbol{\theta}$ are the model parameters. 
We use deep feedforward networks for $f_{\boldsymbol{\theta}}$. Appendix~\ref{appendix_experiments} contains the full details about the experimental setup. 

\subsubsection{Data and Models: Variable-size} 
\label{sec:DataAndModels}

For the variable-size case, we consider a maximal number of possible players $M$ and pad the inputs with zeros for games with less than $M$ players. Hence, we generate a single dataset $\mathcal{D}_{\text{var}} = \left\{ \mathbf{X} \in \mathbb{R}^{G \times M}, \mathbf{P} \in \mathbb{R}^{G \times K} \right\}$. Similarly to the fixed-size dataset, the feature matrix $\mathbf{X}$ contains the normalized weights and $\mathbf{P}$ the corresponding solutions with either $K=M+1$ or $K=M$, with the ground truth output vector again padded with zeros when there are fewer than $M$ players. Hence, we allow for the prediction of up to $M$ players, and we shuffle the data so that players are located at random positions. The Least Core Value (LCV) is not shuffled but stored at the last element of each row (Figure~\ref{fig:model-architecture}).

We produce a single model $g_{\boldsymbol{\eta}}: \mathbb{R}^{n} \to \mathbb{R}^{K}$ that can be used for different number of players $n$, where $\boldsymbol{\eta}$ are the model parameters. The model learns to allocate the joint payoff among at most $M$ players. During prediction time we pad the input with zeros when there are fewer than $M$ players, and we redistribute the payoffs allocated to non-player entries among the players according to their original share of the joint payoff. 

\subsubsection{Training and Evaluation}

During training we minimize the Mean Square Error (MSE) between the true and predicted solutions. For the variable-size models we also include the $M-n$ padded locations so that the model learns not to allocate value to non-player entries. 

\paragraph{Evaluation metrics.}
Given a predicted payoff vector $\hat{\mathbf{p}} = (\hat{p}_1, \hat{p}_2, \dots, \hat{p}_n)$, we consider multiple metrics  for assessing the models' performance. First, we quantify the models' predictive performance via the {{Mean Absolute Error (MAE)}}, defined for each game as $\text{MAE} = \frac{1}{n} \sum_{i = 1}^n  | p_{i} - \hat{p}_{i} |$, where $n$ is the number of players. For the Least-Core there is another natural game theoretic metric. 
The goal of the Least-Core is to minimize the incentives of any subset to abandon the current team and form its own sub-team. Given a suggested payoff vector $\hat{\mathbf{p}}$, the maximal excess $v(C) - \sum_{i \in C} \hat{p}_i$ over all possible coalitions $C$ measures the incentive to abandon the team, and serves as a good measure for the quality of the model. 

\paragraph{Test data.}\label{testdatainfo} We sample weights $\mathbf{w} \sim \text{Beta}(\alpha, \beta)$ with varying parameters for $\alpha$ and $\beta$ to assess our models' ability to generalize to previously unobserved instances their ability generalize to games far outside of the training distribution (full details in Table \ref{tab:parameter_testgames}, Figure  \ref{fig:test_distributions}).

\subsection{Feature Importance Games}\label{sec:methods_XAI}

We perform an experiment to show that neural networks can provide a faster alternative for measuring feature importance at scale. We select a dataset (details in Appendix \ref{sec:exp_details}), train a model, and use those to construct a dataset from features to Shapley values using the SHAP KernelExplainer \cite{lundberg2017unified} Following, we partition our dataset into a train and test, and incrementally change the proportions between the two. For each increment, we train a model for 100 epochs and test it on the remainder of the unseen instances. 

\section{Results}

We present experimental results that allow us to assess how well neural models are able to learn a representation of the various solution concepts. We first describe the predictive performance of
neural networks in the WVG setting, and consider properties of these solutions that make them hard to learn. We then consider the explainable AI domain, and study the performance and sample complexity of Shapley feature importance prediction.

\subsection{Weighted Voting Games: Evaluation}

For our WVG analysis we train a selection of fixed-size neural networks $f_{\boldsymbol{\theta}^{*}}^N$ for each number of players $n \in \{4, 5, \dots, 19, 20 \}$\footnote{A direct computation of the Shapley value requires enumerating through a large list of permutations, which becomes computationally very costly when there are many players. Hence, for games with $n=9$ players or more, we use Monte-Carlo approximations for the Shapley value to obtain the ground-truth solution. See full details in Appendix \ref{sec:approx_shap_values})} 
on $G=5,000$ games each. We also train a single variable-size model $g_{\boldsymbol{\eta^{*}}}$ that is trained on one dataset containing $G=17,500$ games in total, consisting of $2,500$ games for each number of players $n \in \{4, 5, \dots, 9, 10 \}$ games. The data is  padded with zeros to allow for payoff allocation up to $M=20$ players (see details in Section~\ref{modelanddata}). 


\subsubsection{Predictive performance}
\label{l_Sect_pred_perf}

Table~\ref{tab:Summary_pred_performance} shows that the ability to handle games of variable numbers of players comes at the cost of having a lower accuracy. However, even for variable-size models, and even under a significant distribution shift, the errors in predicting all solution concepts are low. We further note that the error in predicting Least-Core based payoffs are generally larger than for the Shapley and Banzhaf power indices. One possible reason is that for these solutions, there are many cases where a small perturbation in the weights or quota results in a large perturbation of the Least-Core solution~\cite{elkind2009computing,zuckerman2012manipulating}.

\begin{table}[ht]
\centering
\caption{\label{tab:Summary_pred_performance} Comparison of predictive performance across test datasets and solution concepts.}
\vspace{2mm}
\resizebox{1\textwidth}{!}{%
\begin{tabular}{lllllllll}
\toprule
& \multicolumn{2}{c}{\textbf{Least core payoffs}} & \multicolumn{2}{c}{\textbf{Least core excess}} & \multicolumn{2}{c}{\textbf{Shapley values}} & \multicolumn{2}{c}{\textbf{Banzhaf indices}} \\\midrule
Dataset & \multicolumn{2}{c}{Mean MAE} & \multicolumn{2}{c}{MAE} & \multicolumn{2}{c}{Mean MAE} & \multicolumn{2}{c}{Mean MAE} \\
 & Fixed & Variable & Fixed & Variable & Fixed & Variable & Fixed & Variable \\ \midrule
In-sample  &  0.030 &  0.043 &  0.015 & 0.034  &  0.019  & 0.022  &  0.018 & 0.028 \\
Out-of-sample & 0.030 & 0.044 & 0.015 & 0.027  & 0.018  & 0.036 & 0.018 & 0.056  \\
Slightly out-of-distribution & 0.029  & 0.028  &  0.015  & 0.050  & 0.018 & 0.019 & 0.017 & 0.018 \\
Moderately out-of-distribution & 0.030  & 0.035 & 0.014  & 0.029 &  0.018 &  0.026 &  0.018 & 0.032  \\
Significantly out-of-distribution & 0.031 & 0.045 & 0.014 & 0.036 & 0.018 & 0.029 & 0.018 & 0.039 \\
\bottomrule
\end{tabular}}
\end{table}
We now highlight several conclusions from our analysis of the results.

\paragraph{Fixed-size models display stable performance across solution concepts.} Our first observation is that the fixed-size neural networks are adept at estimating solutions across the three considered concepts. The average error per player ranges from $0.005$ to $0.087$ (Least Core), $0.004$ to $0.069$ (Shapley values) and $0.012$ to $0.067$ (Banzhaf indices). Table \ref{tab:Predictive_performance_insample} provides a complete summary of our models' in-sample performance.

\begin{figure}[h!t] 
\centering
\begin{subfigure}{.44\textwidth}
    \centering
    \includegraphics[width=1\linewidth]{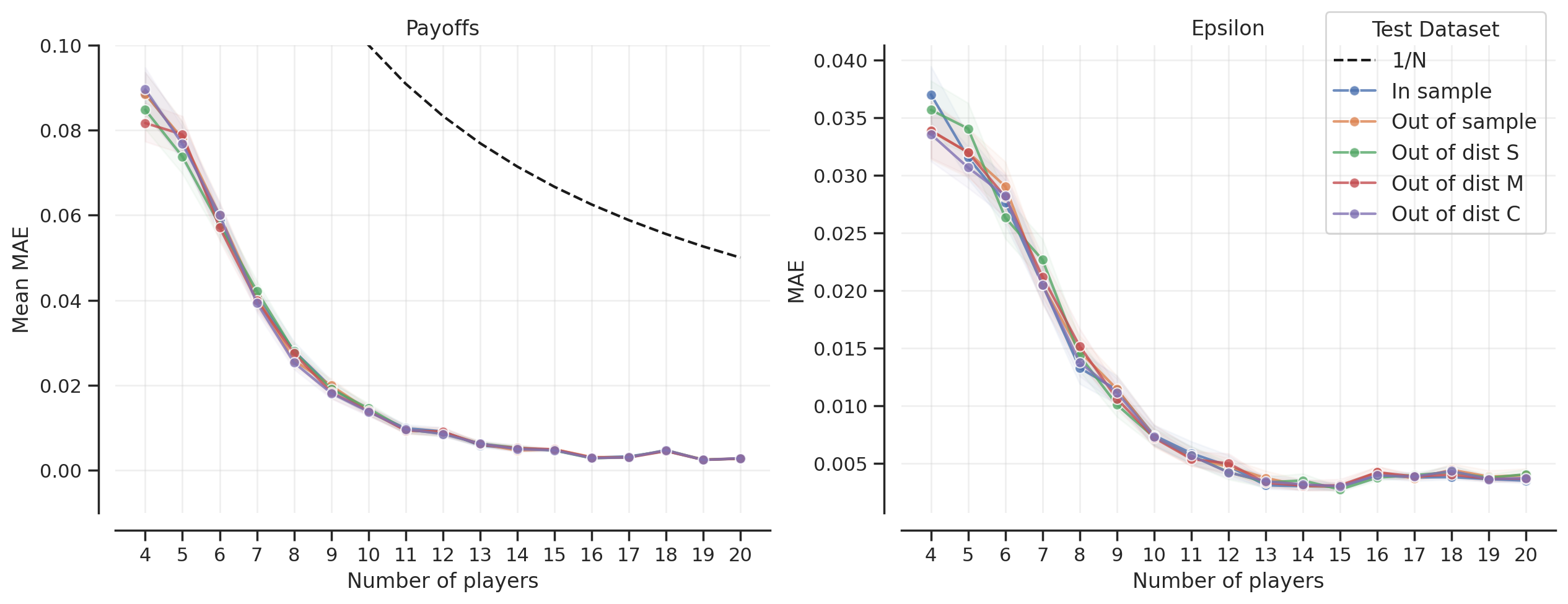}  
    \caption{Least Core}
    \label{SUBFIGURE LABEL 1}
\end{subfigure}
\begin{subfigure}{.26\textwidth}
    \centering
    \includegraphics[width=1\linewidth]{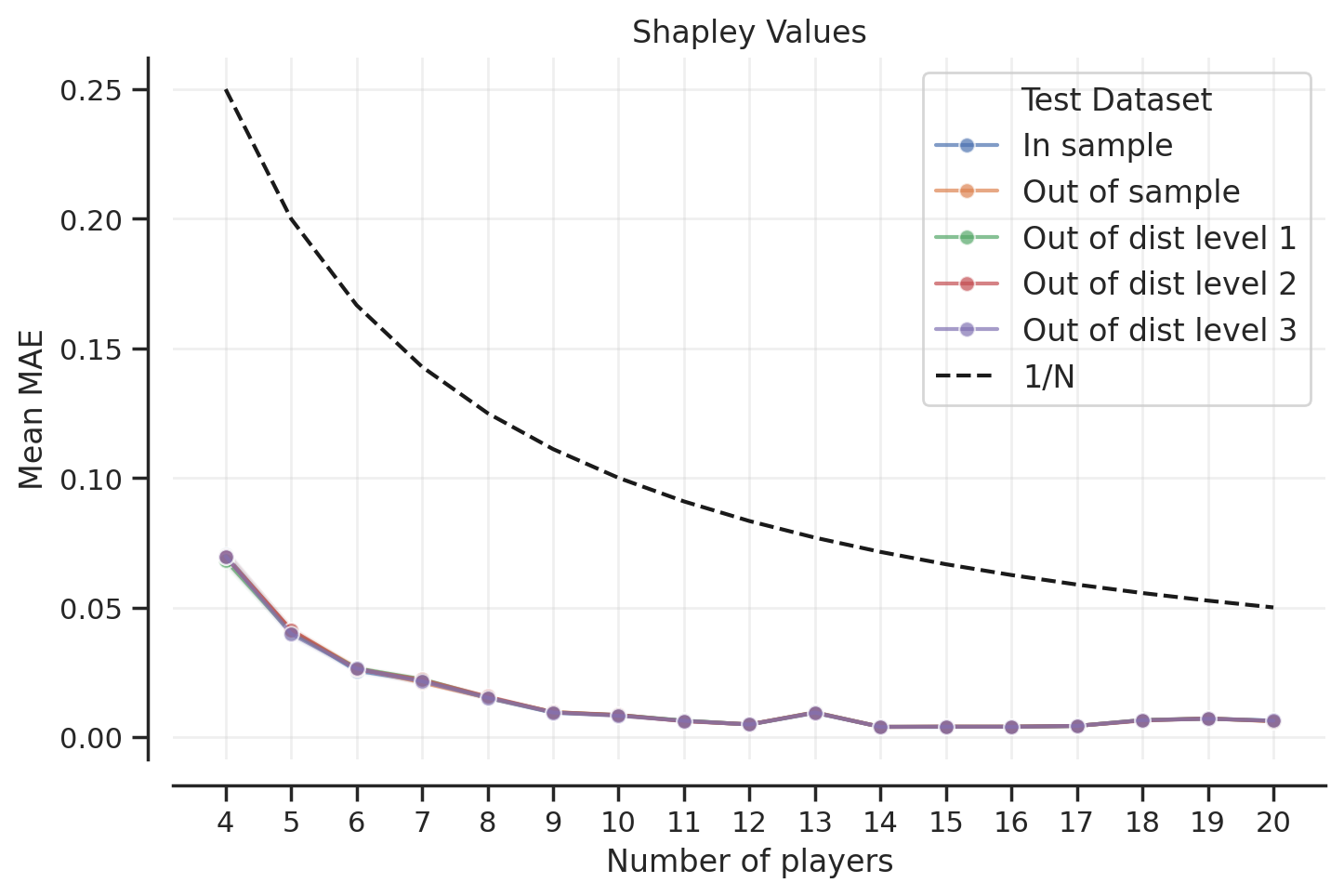}
    \caption{Shapley value}
    \label{SUBFIGURE LABEL 2}
\end{subfigure}
\begin{subfigure}{.26\textwidth}
    \centering
    \includegraphics[width=1\linewidth]{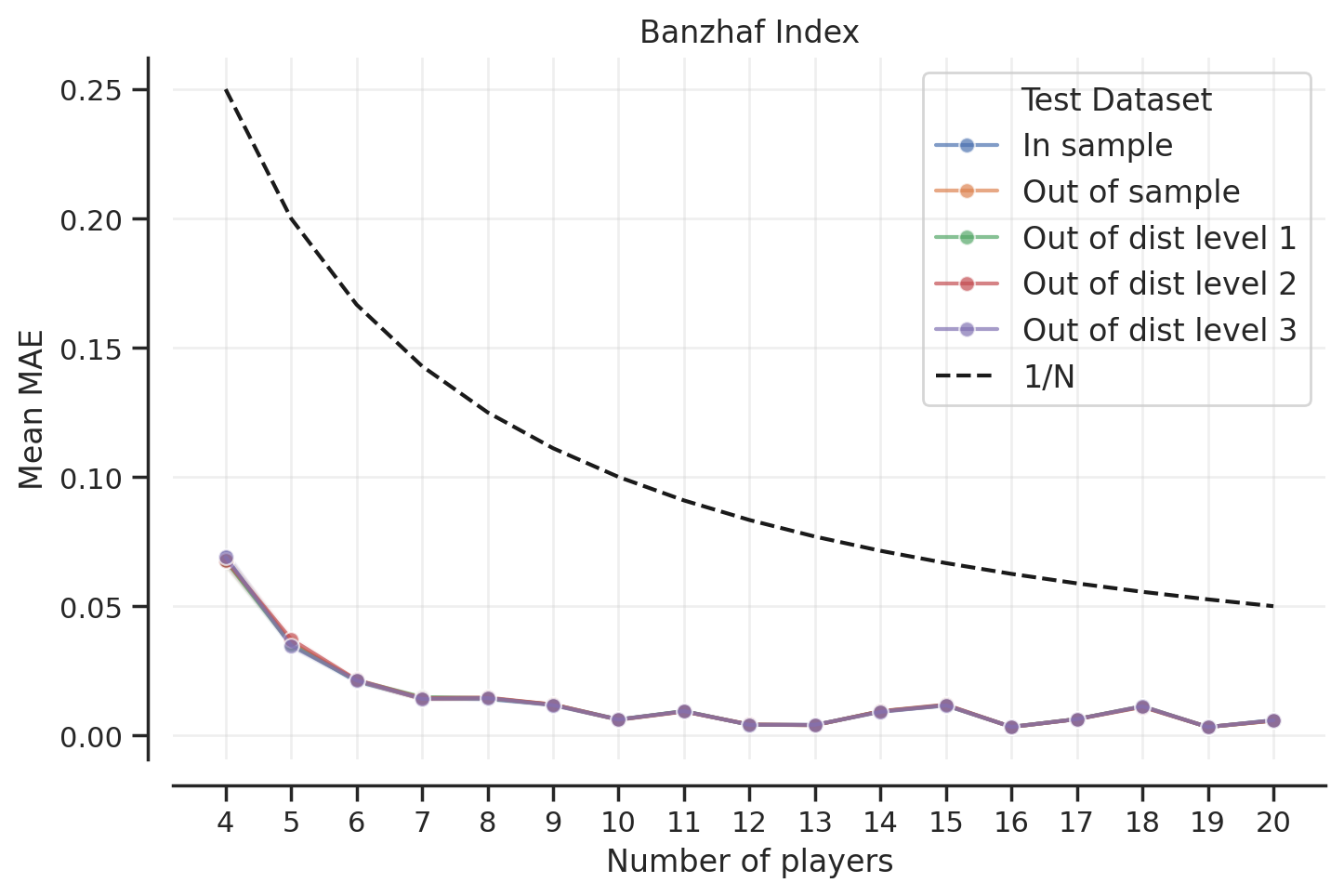}
    \caption{Banzhaf index}
    \label{SUBFIGURE LABEL 3}
\end{subfigure}
\caption{\textbf{Performance fixed-size models.} We evaluate the performance on five test sets with $1000$ samples per $N$ player each. Figures show the MAE and 95 \% confidence intervals for each solution concept. The black dashed line indicates the average payoff per player as a function of $N$.}
\label{fig:Fixed_summary}
\end{figure}

\paragraph{Fixed-size models are robust to shifts in the weight distribution.}
As shown in Figure \ref{fig:Fixed_summary}, the predictive performance (Mean MAE) of the fixed-size models is consistent across test datasets. To account for the natural decrease in the MAE as $N$ increases, we also display the average payoff per player ($1/n$). As expected, the error scales approximately with the average payoff per player. 


\begin{figure}[h!t] 
\centering
\begin{subfigure}{.44\textwidth}
    \centering
    \includegraphics[width=1\linewidth]{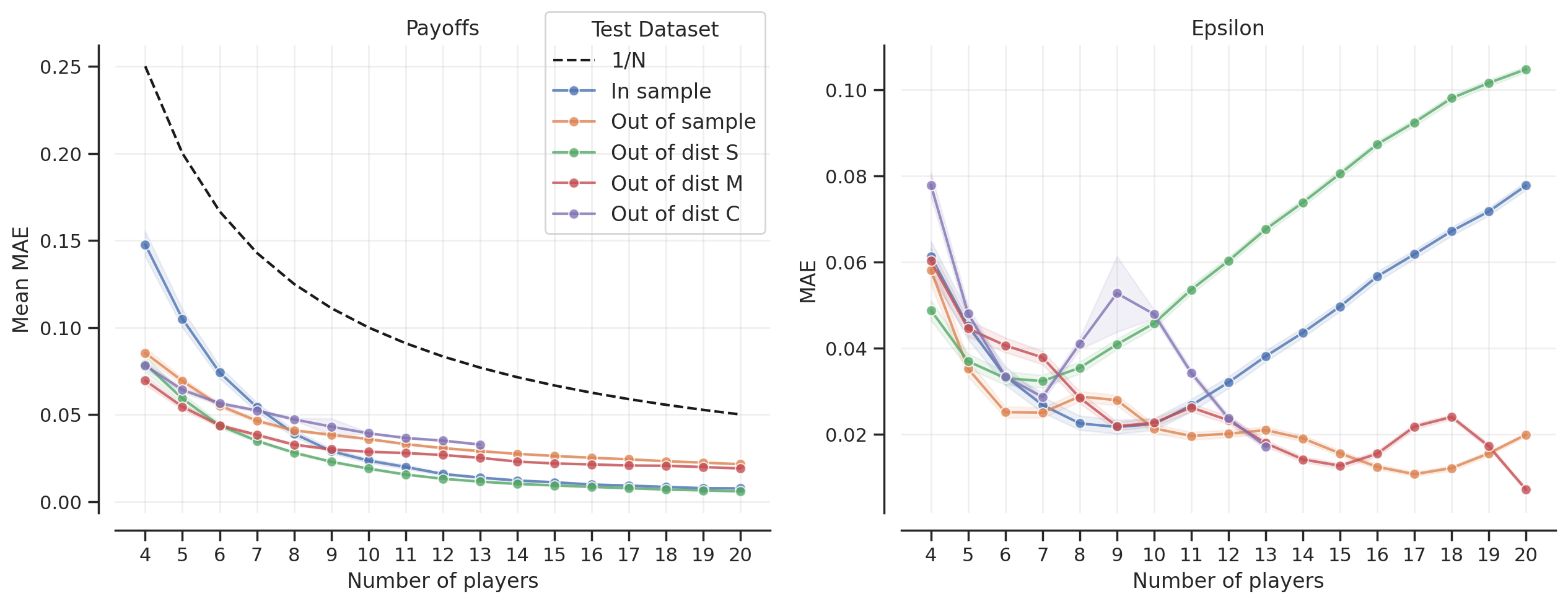}  
    \caption{Least Core}
    \label{SUBFIGURE LABEL 1}
\end{subfigure}
\begin{subfigure}{.26\textwidth}
    \centering
    \includegraphics[width=1\linewidth]{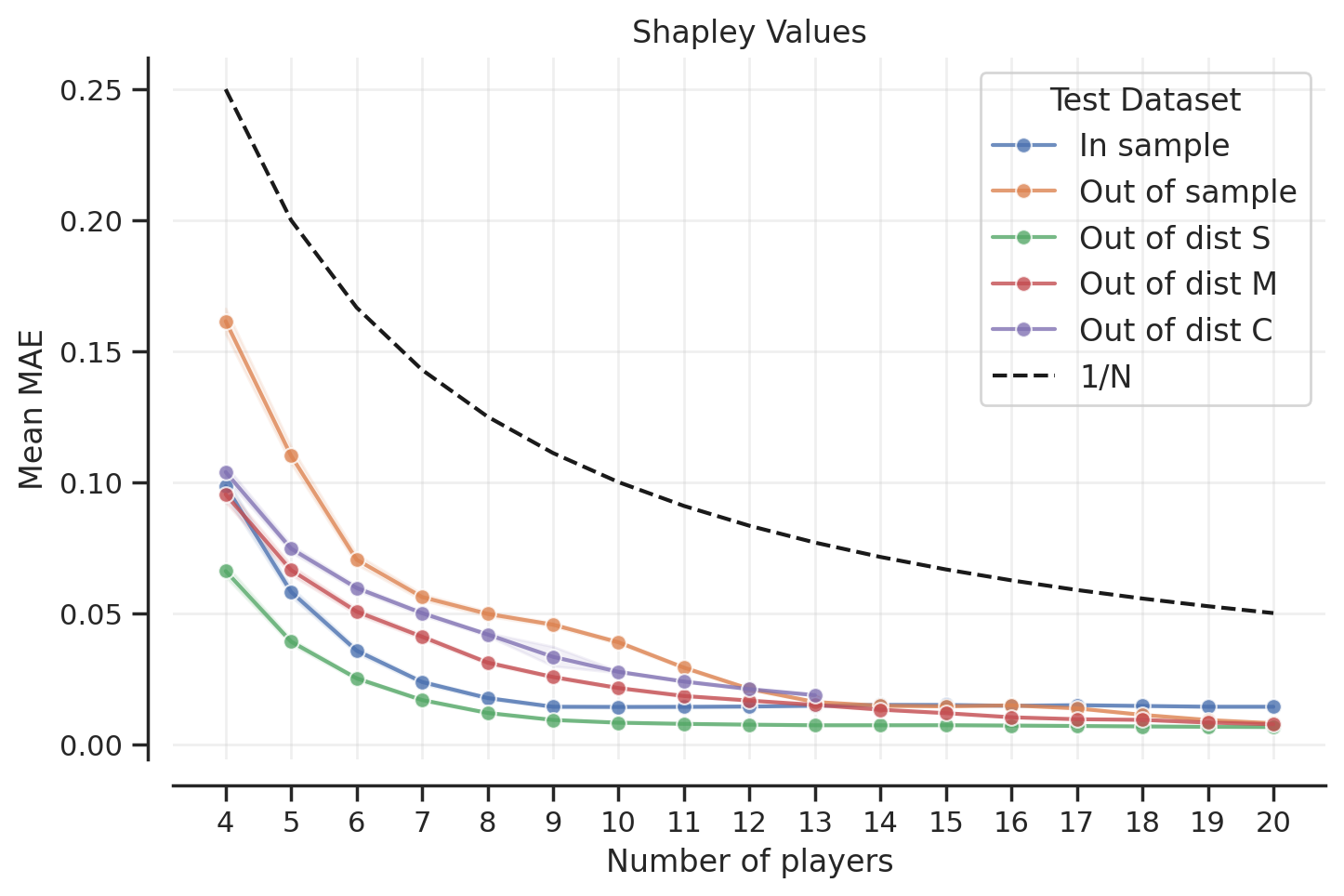}
    \caption{Shapley value}
    \label{SUBFIGURE LABEL 2}
\end{subfigure}
\begin{subfigure}{.26\textwidth}
    \centering
    \includegraphics[width=1\linewidth]{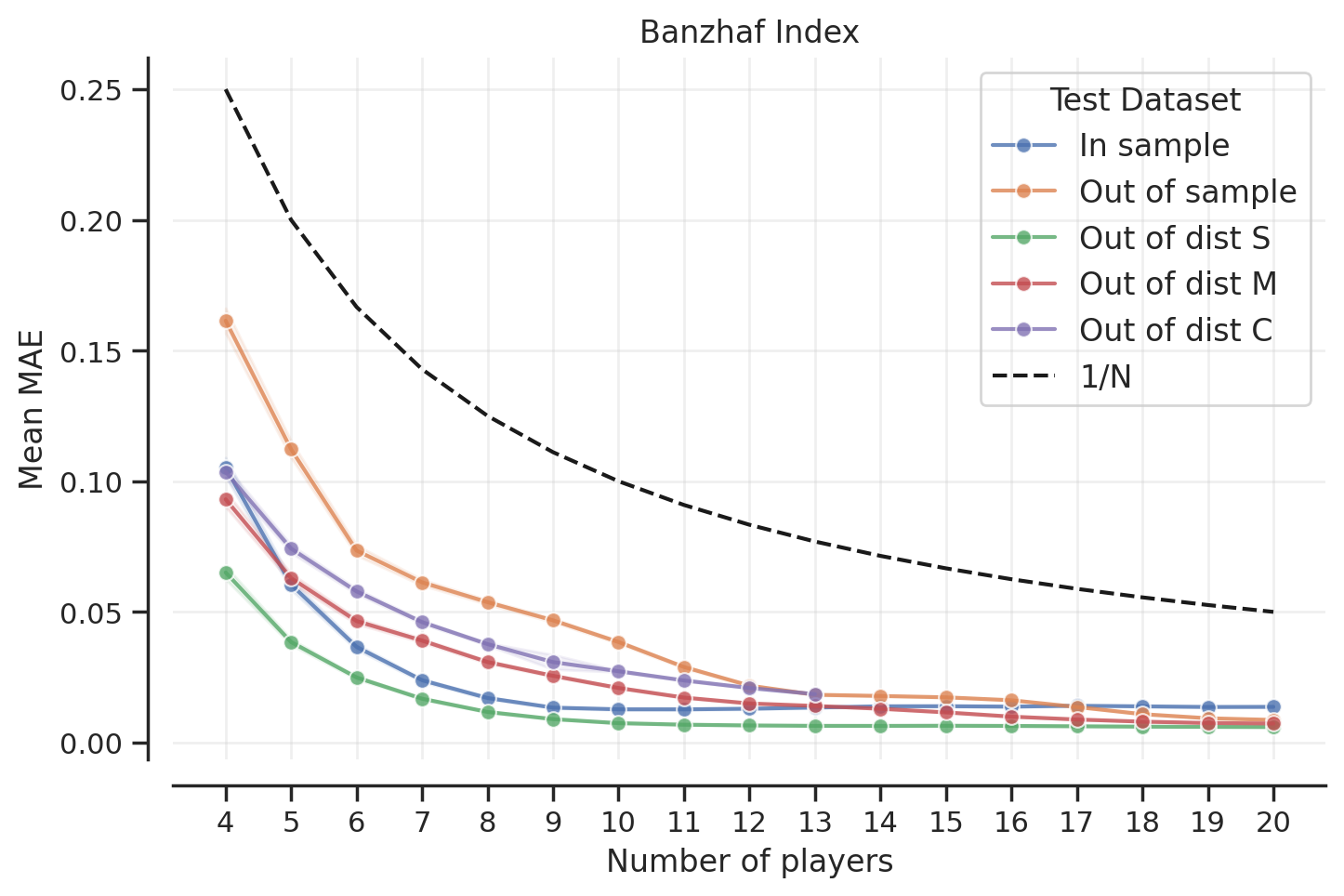}
    \caption{Banzhaf index}
    \label{SUBFIGURE LABEL 3}
\end{subfigure}
\caption{\textbf{Performance variable-size models.} We evaluate the performance on five test sets with 1000 samples per $N$-player game. Figures show the MAE and 95 \% confidence intervals for each solution concept. The black dashed line indicates the average payoff per player as a function of $N$.}
\label{fig:Var_summary}
\end{figure}

\paragraph{Variable-size models are robust to shifts in the weight distribution.} Figure \ref{fig:Var_summary} shows that the variable-size models are also able to generalize outside the training distribution. Across test sets, we observe a stable performance that decays with the number of players, as is expected. For the Least Core, we see that there is an abrupt decrease in performance for the excess beyond 8 players.

\paragraph{Variable-size models generalize to a larger number of players.} Our main objective is to investigate how to leverage machine learning to perform scalable value estimation in large multi-agent settings. To test this, we train the variable-size models on games up $N = 10$ players, and evaluate on $N+1, N+2, \dots , N+10$ players (full details in section \ref{sec:DataAndModels}). Figure \ref{fig:Var_summary} demonstrates that there is no significant decrease in performance for games with more than ten players across solution concepts: variable-size models are able to extrapolate beyond the number of players seen during training. This suggests that there is be valuable information in the small player games such that games of larger sizes can be inferred.



\begin{figure}[h!t] 
\centering
\begin{subfigure}{.44\textwidth}
    \centering
    \includegraphics[width=1\linewidth]{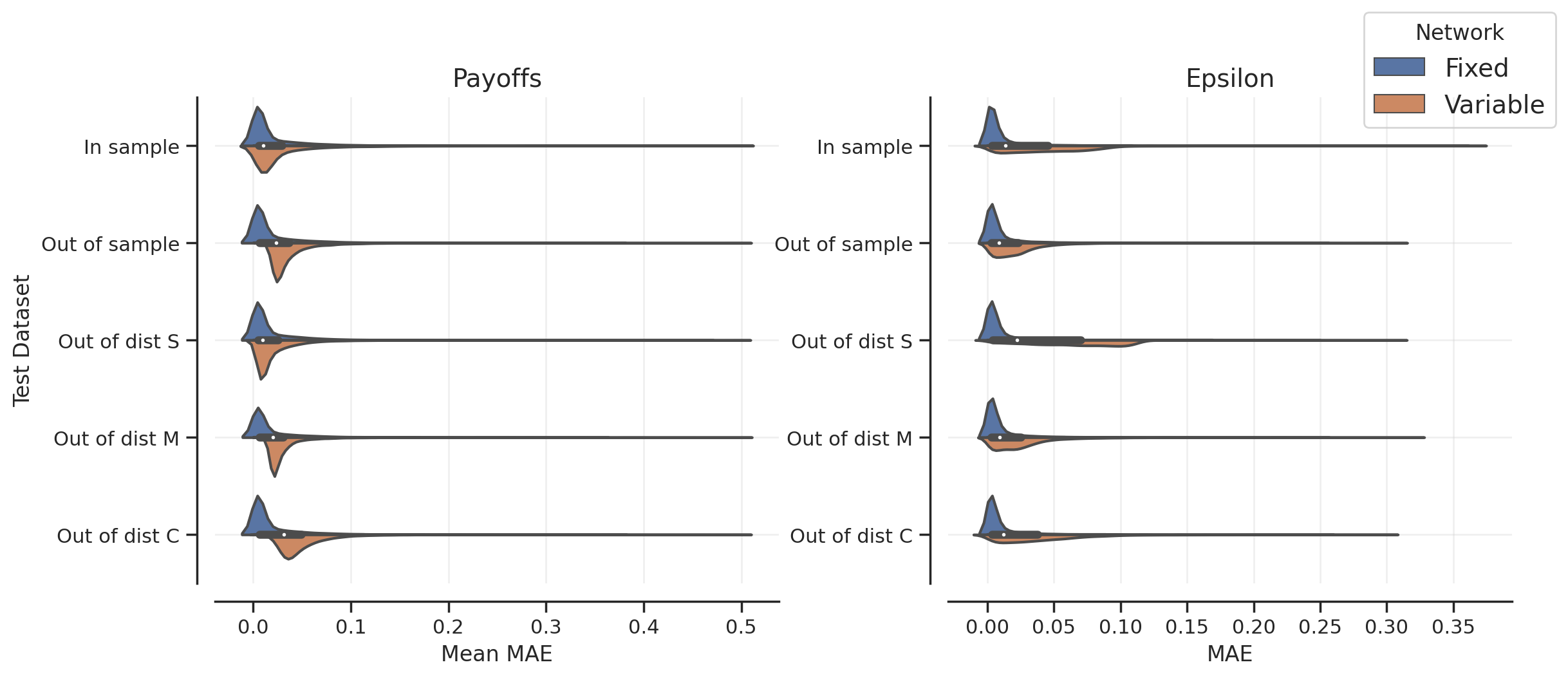}  
    \caption{Least Core}
    \label{SUBFIGURE LABEL 1}
\end{subfigure}
\begin{subfigure}{.26\textwidth}
    \centering
    \includegraphics[width=1\linewidth]{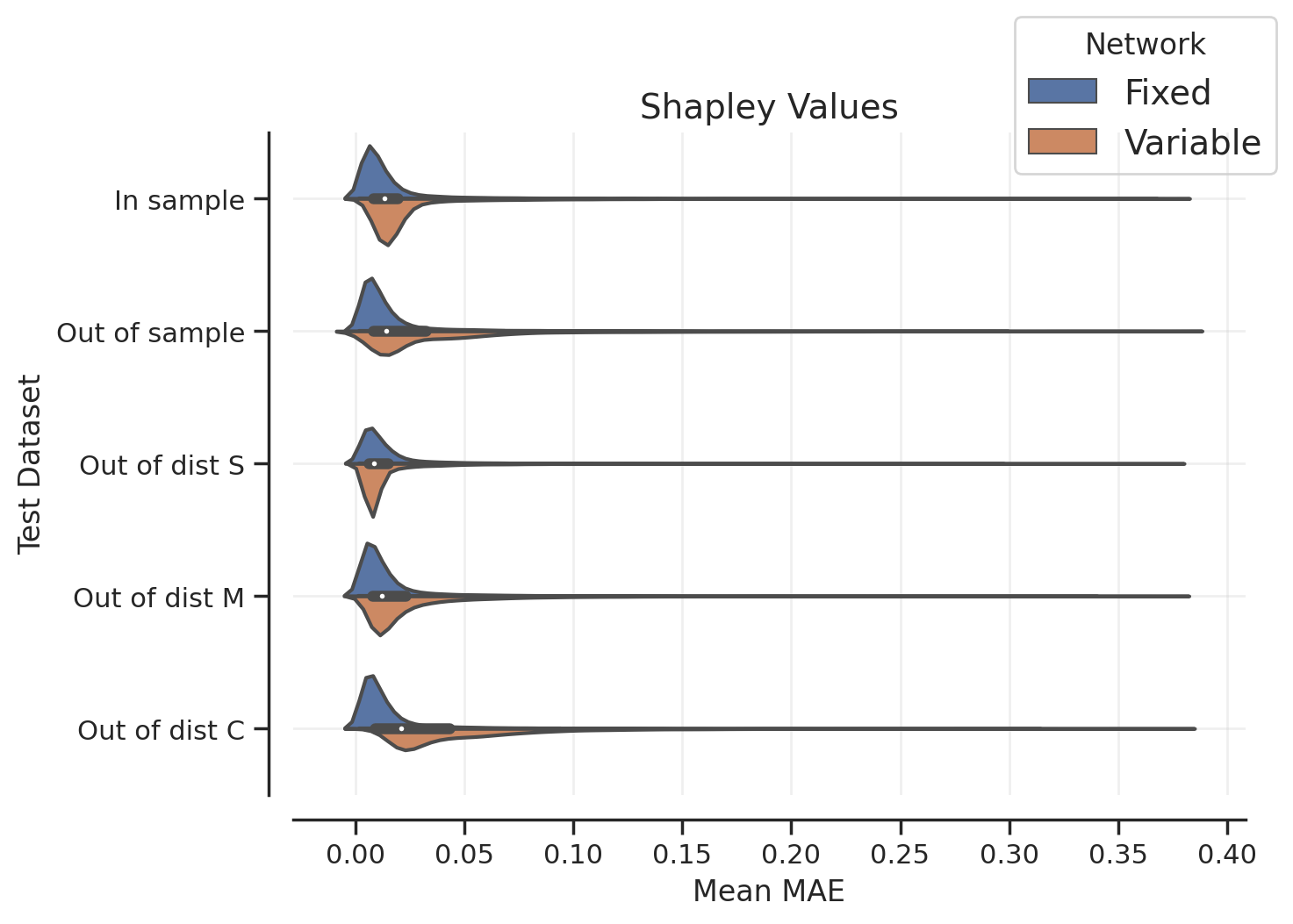}
    \caption{Shapley value}
    \label{SUBFIGURE LABEL 2}
\end{subfigure}
\begin{subfigure}{.26\textwidth}
    \centering
    \includegraphics[width=1\linewidth]{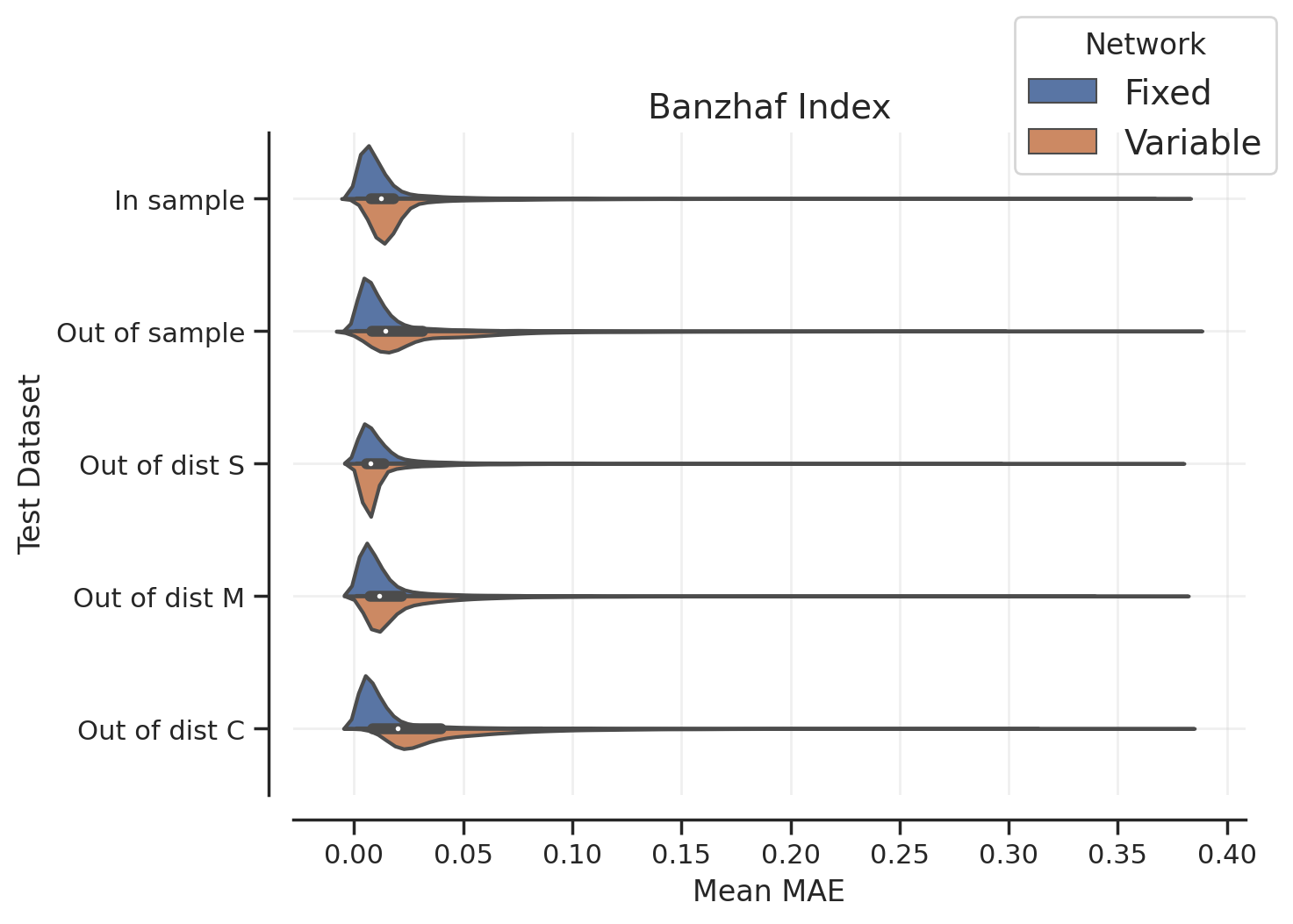}
    \caption{Banzhaf index}
    \label{SUBFIGURE LABEL 3}
\end{subfigure}
\caption{\textbf{Comparing overall performance fixed-size vs. variable-size networks.}}
\label{fig:Comparison_summary}
\end{figure}

\paragraph{Fixed-size models outperform variable-size models.} Table \ref{tab:Summary_pred_performance} contains the Mean MAEs across solution concepts and test sets. Corresponding error distributions are showed in Figure \ref{fig:Comparison_summary}. From these, we conclude that in almost all settings the fixed-size networks outperform the variable-size networks. Variable-sized networks tend to have larger errors for all predicted variables and display more variance in their predictions.

\subsubsection{Discontinuities in the Solution Space}\label{sec:capture_disc}
The function mapping from game parameters to solutions contains discontinuities. Discontinuous jumps emerge from the players' interdependence and the effect of the quota and are difficult for a model to learn. We analyze two examples that demonstrate how our models respond to such transitions.



\paragraph{Solution concepts are step-wise constant functions which are difficult to capture.} Consider taking a WVG and changing the weight of a player. This only changes the game theoretic solutions when it restructures the subset of winning coalitions. Thus, the function outputs the same value until the weight reaches a certain threshold where the structures of winning coalitions change, at which point the solution can change drastically. Learning these kind of functions is difficult, as the error around the discontinuity point is often large. 

\begin{figure}[h!t]
     \centering
     \begin{subfigure}[b]{0.49\textwidth}
         \centering
         \includegraphics[width=\textwidth]{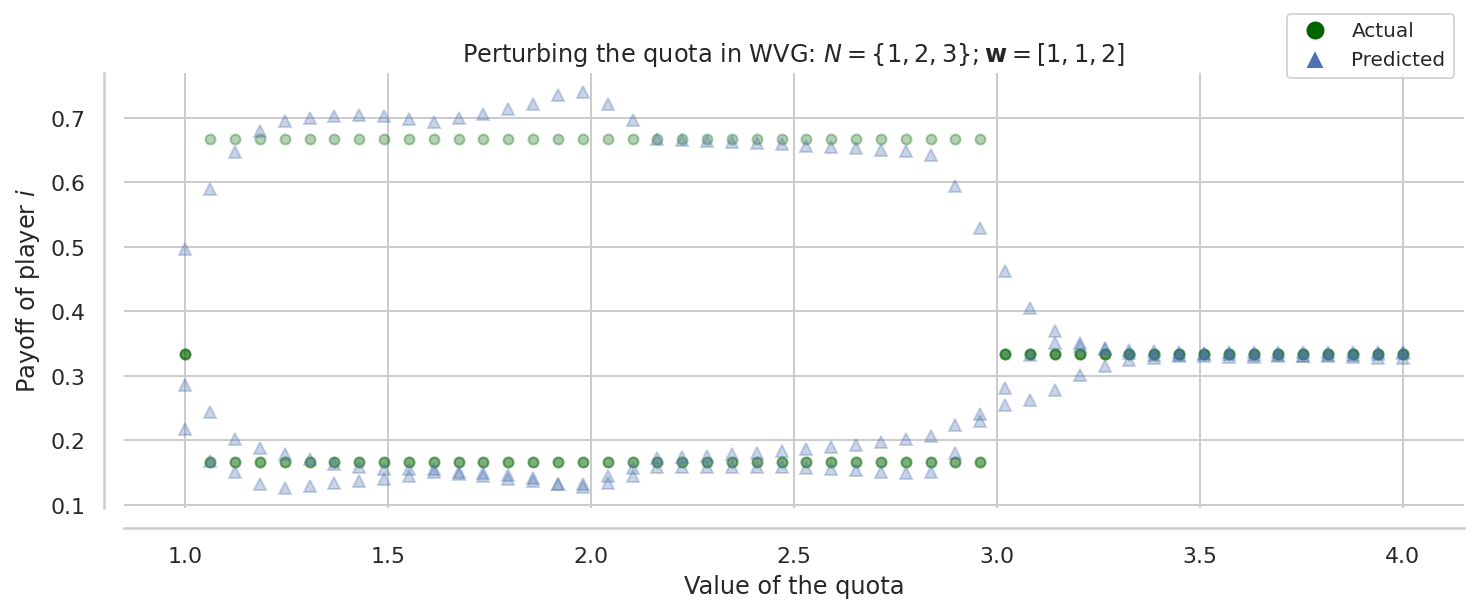}
     \end{subfigure}
     \hfill
     \begin{subfigure}[b]{0.49\textwidth}
         \centering
         \includegraphics[width=\textwidth]{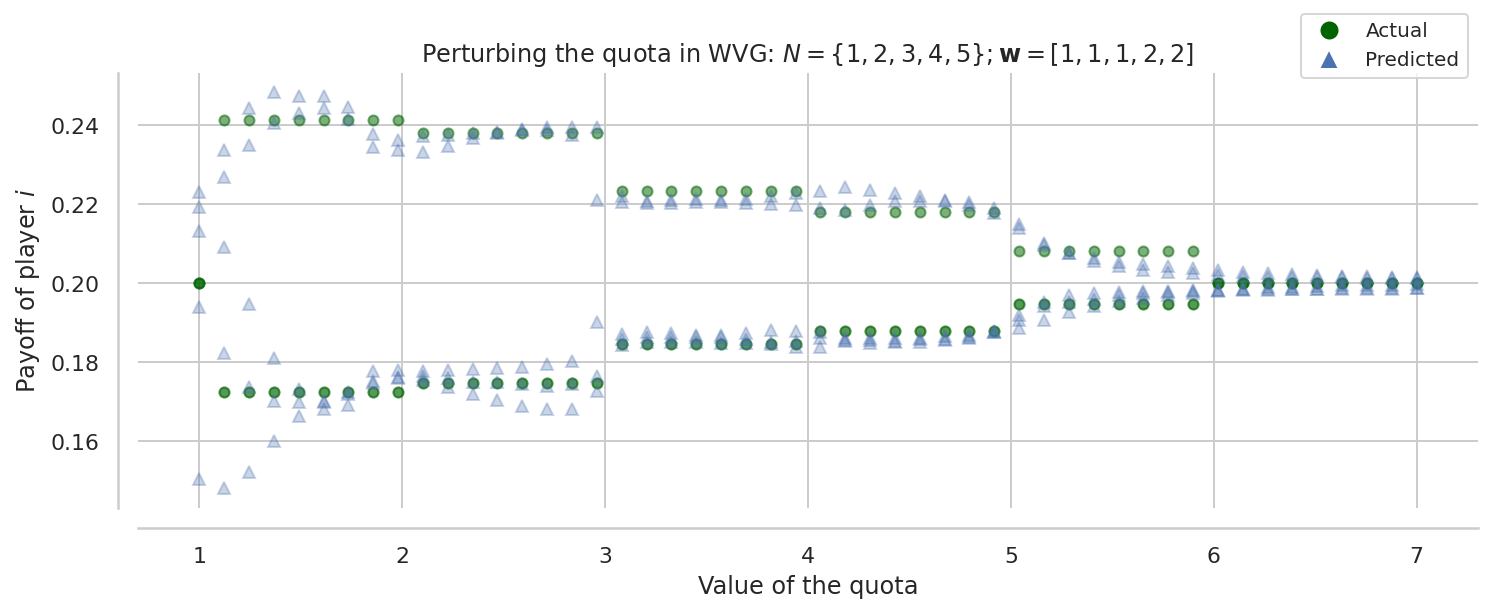}
     \end{subfigure}
        \caption{\textbf{Capturing step-wise jumps.} Individual actual payoffs (green dots) together with the model predictions (blue triangles) as the value of quota increases by $\varepsilon = 0.1$ increments (full description in Appendix \ref{sec:stepwise_desc}).}
        \label{fig:perturb_quota}
\end{figure}

Our analysis examines an $n$ player game with a fixed weight vector $\mathbf{w} = (w_1, w_2, \dots, w_n)$ and considers an array of quotas $\mathbf{q} = (q_1, q_2, \dots, q_K) = \left( \text{min}(\mathbf{w}), \text{min}(\mathbf{w}) + \epsilon , \dots, \sum_{i = 1}^n w_i \right)$, where $\epsilon = 0.1$. We solve the game for each combination of the fixed weights and the changing quota in $\mathbf{q}$ to obtain a matrix $\mathbf{P}$. Figure~\ref{fig:perturb_quota} shows the fixed-size model predictions for two selected WVGs. The models capture the overall effect of the changes to the quota, but do poorly close to the discontinuity point (where the ground truth solution incurs a large change).

\subsection{Speeding Up Feature Importance Calculations By Predicting Shapley Values}
\begin{figure}[h!t] 
     \centering
     \begin{subfigure}[b]{0.49\textwidth}
         \centering
         \includegraphics[width=1\textwidth]{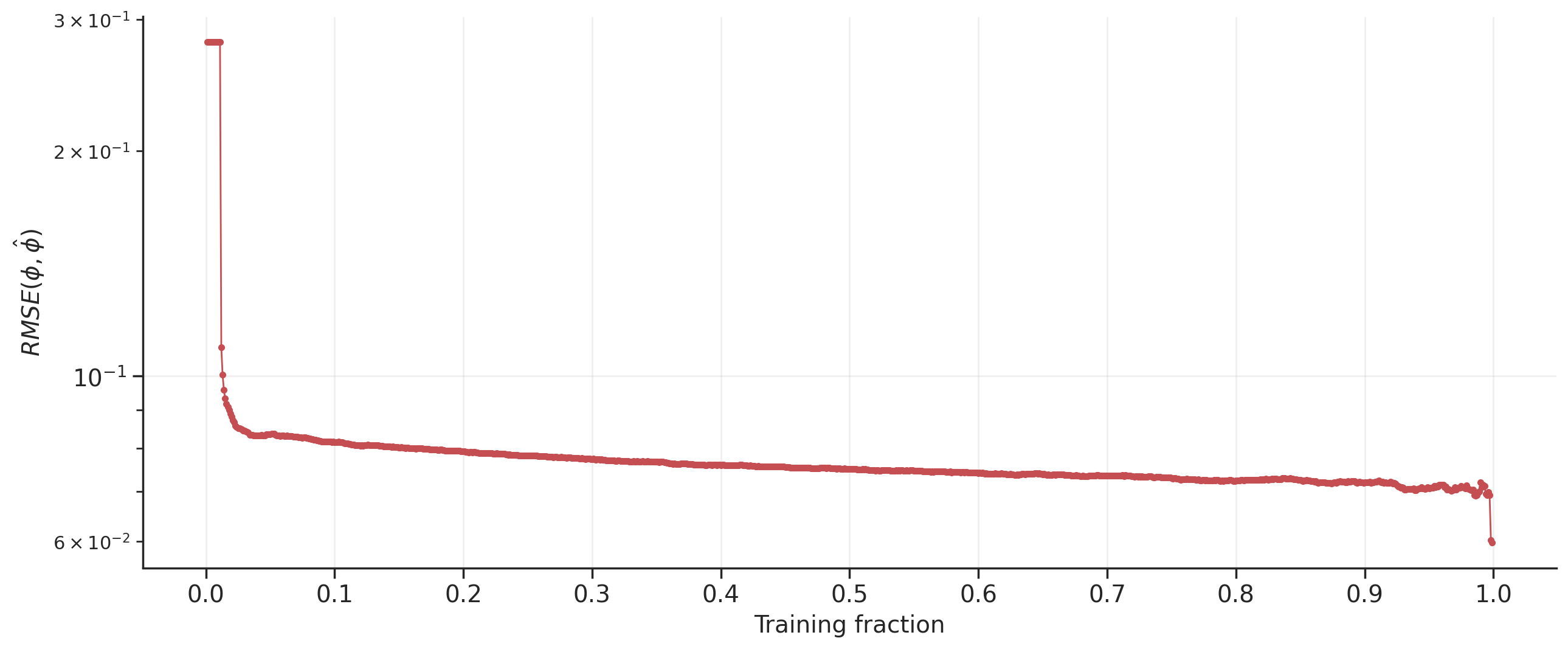}
     \end{subfigure}
     \hfill
     \begin{subfigure}[b]{0.49\textwidth}
         \centering
         \includegraphics[width=1\textwidth]{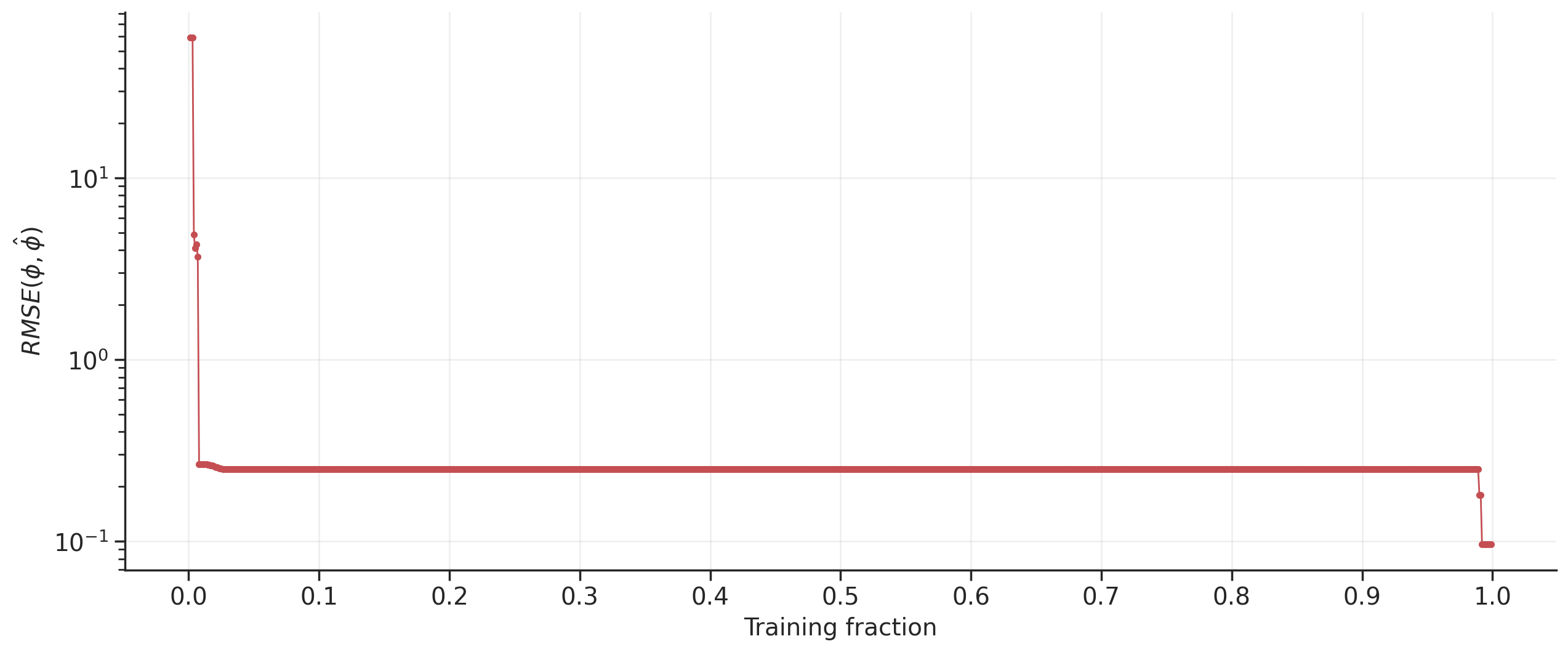}
     \end{subfigure}
        \caption{\textbf{Root Mean Squared Error as a function of training fraction.} The RMSE between the actual and predicted Shapley values scales approximately as a power law with the training fraction. \textit{Left}: UCI Banking dataset \textit{Right}: Melbourne Housing dataset.}
        \label{fig:Explainable_ai_shap}
\end{figure}

We examine the sample efficiency and performance of neural networks trained to predict Shapley values on sets of features. We consider two well-known datasets: the classical UCI Bank Marketing dataset \cite{moro2014data} (17 features, 11,162 observations) and the Melbourne Housing dataset \cite{pino2018melbourne} (21 feature, 34,857 observations). Full details in Appendix \ref{sec:explainable_ai}). 
Overall, we find that neural networks achieve high performance on the test set with very few training samples. Figure \ref{fig:Explainable_ai_shap} displays the models’ performance (in RMSE) as a function of samples available for training. We observe that the error decays approximately exponentially with the proportion of data used for training on both the Banking and Melbourne datasets. A model trained on just 1 percent of the Melbourne dataset has an RMSE of 19.72 on the resultant test set. Increasing the number of training samples to 3 percent results in a RMSE of less than 0.07, a 99.6 percent decrease in error. We observe a similar trend for the Banking dataset: the RMSE decreases with 93.8 percent (from 4.10 to 0.25) when the available training data increases from 0.5 to 4 percent. These results highlight the strength of this approach: the computational cost of training the Shapley prediction network is very small as compared to the speedup obtained on the vast majority of the data (and any subsequent instances analyzed later). See Appendix  \ref{sec:in_depth_shap_exp} for a more extensive analysis.


\section{Discussion and Conclusion}
We considered ``neural payoff machines'', models that take in a representation of a cooperative game and solve it to produce a distribution of the joint gains of a team amongst its members. Our analysis focused on two concise representations of a characteristic function: Weighted voting games, and feature importance games for explainable AI. Our analysis shows that neural models capture cooperative game solutions well, and can generalize well outside the training distribution, even extrapolating to more players than in the training set. 

We saw that the Least core is a harder solution concept to learn than the Shapley value and Banzhaf index. Potentially, the fact that the Least core is complex and can only be computed by solving multiple Linear Programs makes a hard concept to grasp from examples. 

The methods we discussed can drive better analysis of decision making bodies and for approximating feature importance, but such methods can also be applied to other classes of cooperative games, so long as one can generate data consisting of game samples and their solutions. For instance, the same approach can be used for other cooperative representations, such as graph-based games~\cite{deng1994complexity,aziz2009power,bachrach2008power} or set-based representations~\cite{ohta2008anonymity}. The direct applications that we have considered in this paper are also intrinsically valuable on their own, as we allow speeding up explainable AI analysis and political influence estimation. Some questions remain open for future work. Can similar techniques be used for non-cooperative games or for other types of cooperative games? Can this approach be applied to other solutions such as the Kernel or Nucleolous~\cite{chalkiadakis2011computational}? Finally, are there better neural network designs, such as graph neural networks or transformers, that can exploit invariances or equivariances in the game's representation or be better equipped to deal with sharp discontinuities in the solutions? 

\bibliographystyle{abbrvnat}
\newpage
\bibliography{references}

\newpage
\section*{Appendix}
\appendix
\counterwithin{figure}{section}

\section{Notation}
The following notation is used throughout this paper:

\begin{itemizeReduced}
    \item $n$ the number of players in the game
    \item $G$ the number of games
    \item $M$ the maximum sequence length in variable feedforward networks
    \item $q$ the quota, threshold that determines when a coalition can complete the goal or task
    \item $w_i$ the weight of a player $i$
    \item $x_i = w_i / q$, the normalized weight of a player $i$
    \item $N$ the grand coalition
    \item $C$ a coalition (subset of players)
    \item $C^{\text{min}}$ the minimal set of winning coalitions: the coalitions where each player is pivotal
    \item $\mathbf{X}$ feature matrix
    \item $\mathbf{P}$ solution matrix 
    \item $p_i$ the allocated reward or payoff to a player $i$ 
    \item $\varepsilon$ the least core value (LCV), i.e. joint payoff to sacrifice such that there exists a feasible payoff vector in the least core
\end{itemizeReduced}

\section{Motivating The Use Of The Shapley Value To Analyze Influence In Weighted Voting Games}\label{l_sect_shap_motivation}

We provide an intuitive example motivating the use of Shapley values for measuring the true influence of participants, in weighted voting games or political power in decision making bodies.

Consider a parliament with 100 seats, and imagine that following the elections we have two large parties of 49 seats each, and one tiny party who has only managed to get 2 seats. However, note that having a majority means controlling a majority of the seats, i.e. more than 50 seats. This means that neither of the two big parties has a majority on its own (and of course the small party hasn't got a majority on its own). 

Hence, to achieve a required majority and form a government or pass a bill, the parties have to work in teams. Clearly, a coalition consisting of the two big parties has a majority, as together they have 49+49=98 seats out of the total of 100. However, a coalition consisting of any of the two big parties and the one small party also has 49+2=51 seats, which is also a majority.

Hence, in the setting discussed above, {\it any} two parties form a winning coalition that has a majority and is able to form a government or pass a bill. This can be modelled as a weighted voting game, where the weights are $(w_1=49, w_2=49, w_3=2)$ and the quota is $q=50$.

Intuitively, as any two parties or more are a winning team and as any single party is a losing team, we might say that the weights are irrelevant. Although the big parties each have almost 25 times the seats as the small party, they all are symmetric in their opportunities to form a winning coalition, and one could claim they actually have equal political power. Clearly, the small party is in a very strong negotiation position, and could demand for instance to control a significant proportion of the joint budget.

In the above weighted voting game, the Shapley value of all agents would be equal (each getting a value of $\frac{1}{3}$), as it considers the marginal contributions of the parties over all permutations of players. Due to the symmetry between the parties, they have identical marginal contributions.

The above specific example illustrates some of the principles behind the fairness axioms that the Shapley value exhibits, and illustrates its importance for measuring influence in decision making bodies.

\section{Algorithm For Generating Least Core Solutions} \label{sec:algorithm_leastcore}

The least core solutions can be generated by considering all possible coalitions. A simple improvement, however, is to use the minimal set of winning coalitions \cite{deegan1978new} instead. In this section, we describe the idea behind this approach and the algorithm used in this paper.

Let $N \equiv \{1, 2, \ldots, n\}$ denote the set of all $N \in \mathbb{Z}_{\geq 0}$ players, let $q \in \mathbb{R}_{\geq 0}$ denote the quota and let $w_i \in \mathbb{R}_{\geq 0}$ denote the weights for player $i$, for all $i \in N$. The characteristic function is given by $v(C) \in [0, 1]$, for coalitions $C \subseteq N$.

The naive approach to find the least core solutions is by considering all winning coalitions $C^{\text{win}}$ defined as
\begin{align}
C^{\text{win}}\equiv \left\{C \subseteq N \,\,\middle|\,\, C \neq \varnothing \text{ and }\sum_{i \in C} w_i \geq q\right\}
\end{align}
and to find a least core solution by solving LP-problem \ref{alg:lp_naive}.
\begin{figure}[ht]
    \vspace{2mm}
    \begin{subfigure}[b]{0.49\textwidth}
    \begin{align*}
        \min_{p_1, p_2, \ldots, p_n, \varepsilon} & &\varepsilon \\
        \text{s.t.} & & \sum_{i \in C} p_i &\geq 1 - \varepsilon & \forall\, C &\in  C^{\text{win}}\\
            & & \sum_{i \in N} p_i &= 1 \\
            & & p_i &\geq 0 & \forall\, i &\in N
    \end{align*}
    \caption{Naive approach}
    \label{alg:lp_naive}
    \end{subfigure}
    \hfill
    \begin{subfigure}[b]{0.49\textwidth}
    \begin{align*}
        \min_{p_1, p_2, \ldots, p_n, \varepsilon} & &\varepsilon \\
        \text{s.t.} & & \sum_{i \in C} p_i &\geq 1 - \varepsilon & \forall\, C &\in  C^{\min}\\
            & & \sum_{i \in N} p_i &= 1 \\
            & & p_i &\geq 0 & \forall\, i &\in N
    \end{align*}
    \caption{Minimal approach}
    \label{alg:lp_efficient}
    \end{subfigure}
    \caption{Algorithms for finding least core solutions.}
    \label{alg:lp_least_core}
\end{figure}
In the LP-problem \eqref{alg:lp_naive}, all winning coalitions are considered, Inside those winning
coalitions there exists a subset, that could branch out and form a smaller coalition.
 This yields a higher payoff for the players in the sub-coalition and is hence a Pareto improvement for the "decision-makers". Therefore, we can instead only consider the subset of \textit{minimal winning} coalitions, given by
\begin{align}
C^{\text{min}}\equiv \left\{C \subseteq N \,\,\middle|\,\, C \neq \varnothing, \sum_{i \in S} w_i \geq q, \text{ and, for all }j \in C, \sum_{i \in C \setminus \{j\}} w_i < q\right\}
\end{align}
and solve the more efficient LP-problem \ref{alg:lp_efficient}.

\section{Experimental details}\label{sec:exp_details}

In this section, we elaborate on the experimental procedures, model architectures and hyperparameters used in our experiments. 

\paragraph{Code and notebook.} Following the guidelines, we provide python code to reproduce our results, a pedagogical notebook (called CooperativeGameTheoryPrimer.ipynb) that introduces the most important game theory formalisms and the necessary data in the accompanied zip file.

\subsection{Weighted Voting Games}
\label{appendix_experiments}

\paragraph{Optimization procedure fixed-size feedforward networks.} For each $N$-player dataset $\mathcal{D}^N_{\text{fixed}}$ we perform $R=100$ independent runs with a maximum of $E=6000$ epochs and an early-stopping criteria which breaks a run if, after a baseline of 500 epochs, the validation loss does not improve for 75 consecutive epochs. For each run, 70 percent is allocated for training and 30 percent for validation. After training, we select the model with the best validation loss. The selected models are stored and used for evaluation. In all experiments we use the Adam \cite{kingma2014adam} optimizer, ReLU \cite{nair2010rectified} activation functions for the hidden layers, Softmax functions for output layer of the payoffs and a Sigmoid for the output layer of epsilon.

\paragraph{Optimization procedure var-size feedforward network.} We perform $R=100$ independent runs for one variable dataset $\mathcal{D}_{\text{var}}$ with a maximum of $E=15000$ number of epochs and an early-stopping criteria which breaks a run if, after a baseline of 500 epochs, the validation loss does not improve for 75 consecutive epochs. For each run, 70 percent is allocated for training and 30 percent for validation. After training, we select the model with the best validation loss, which is stored and used for evaluation. In all experiments we use the Adam \cite{kingma2014adam} optimizer, ReLU \cite{nair2010rectified} activation functions for the hidden layers, Softmax functions for output layer of the payoffs and a Sigmoid for the output layer of epsilon.

\begin{figure}[h!t]
    \centering
    \includegraphics[width=0.9\textwidth]{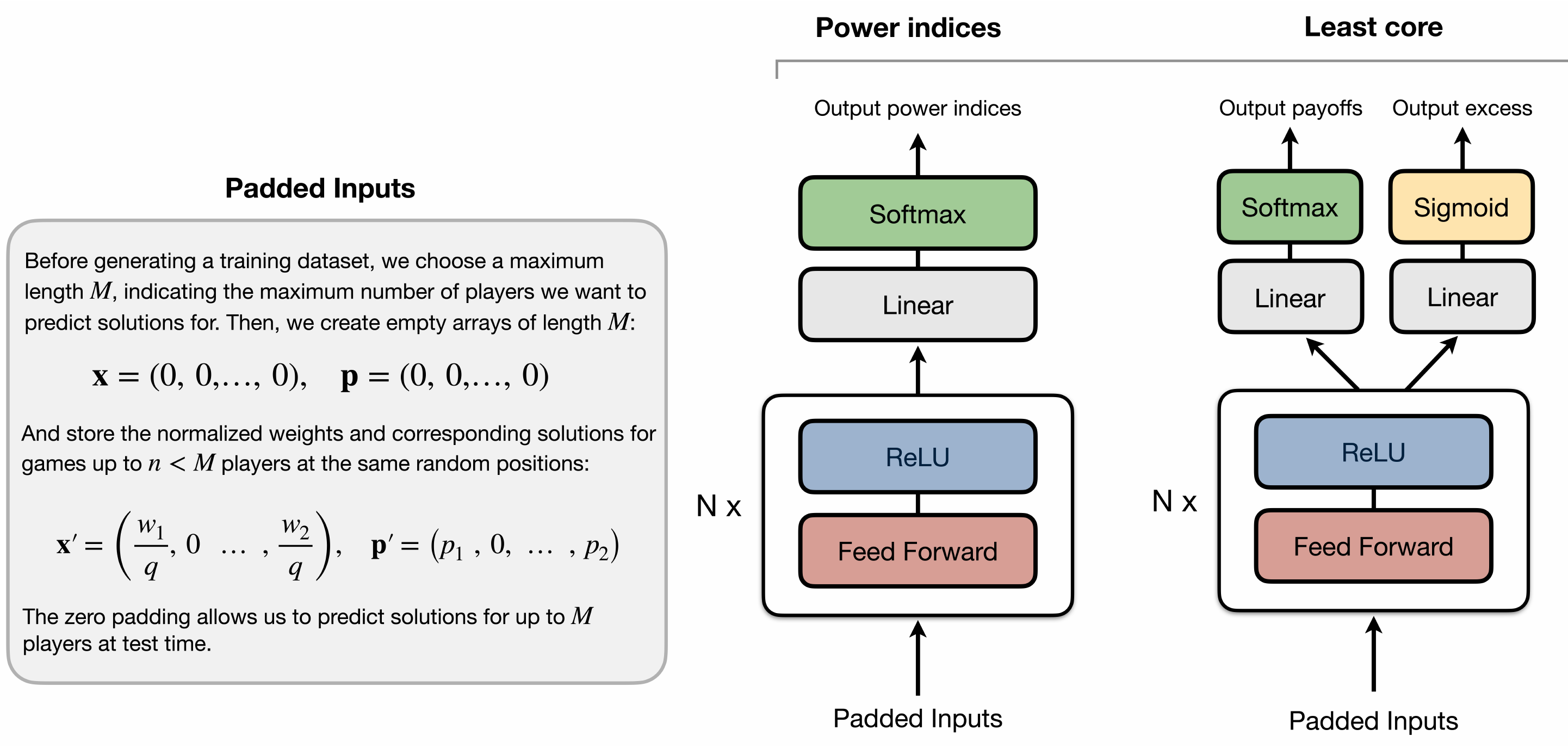}
    \caption{\textbf{Neural architectures for the variable-size game predictions.} We add zero-padding to allow one model to predict solutions of games of variable sizes. We choose a maximum length $M=20$, and add zero-padding to games with less than $M$ players. Full details on our data in Section \ref{sec:DataAndModels}}
    \label{fig:model-architecture}
\end{figure}

\paragraph{Hyperparameters.} All results are generated with multi-layer perceptrons (feedforward networks) of varying hidden-layer size and width. Further hyperparameters are the dropout rate, weight decay in the Adam optimizer, and the learning rate. Our standard model architectures are depicted in ~Figure \ref{fig:model-architecture} Right.

\begin{figure}[h!t]
     \centering
     \begin{subfigure}[b]{0.48\textwidth}
         \centering
         \includegraphics[width=1\textwidth]{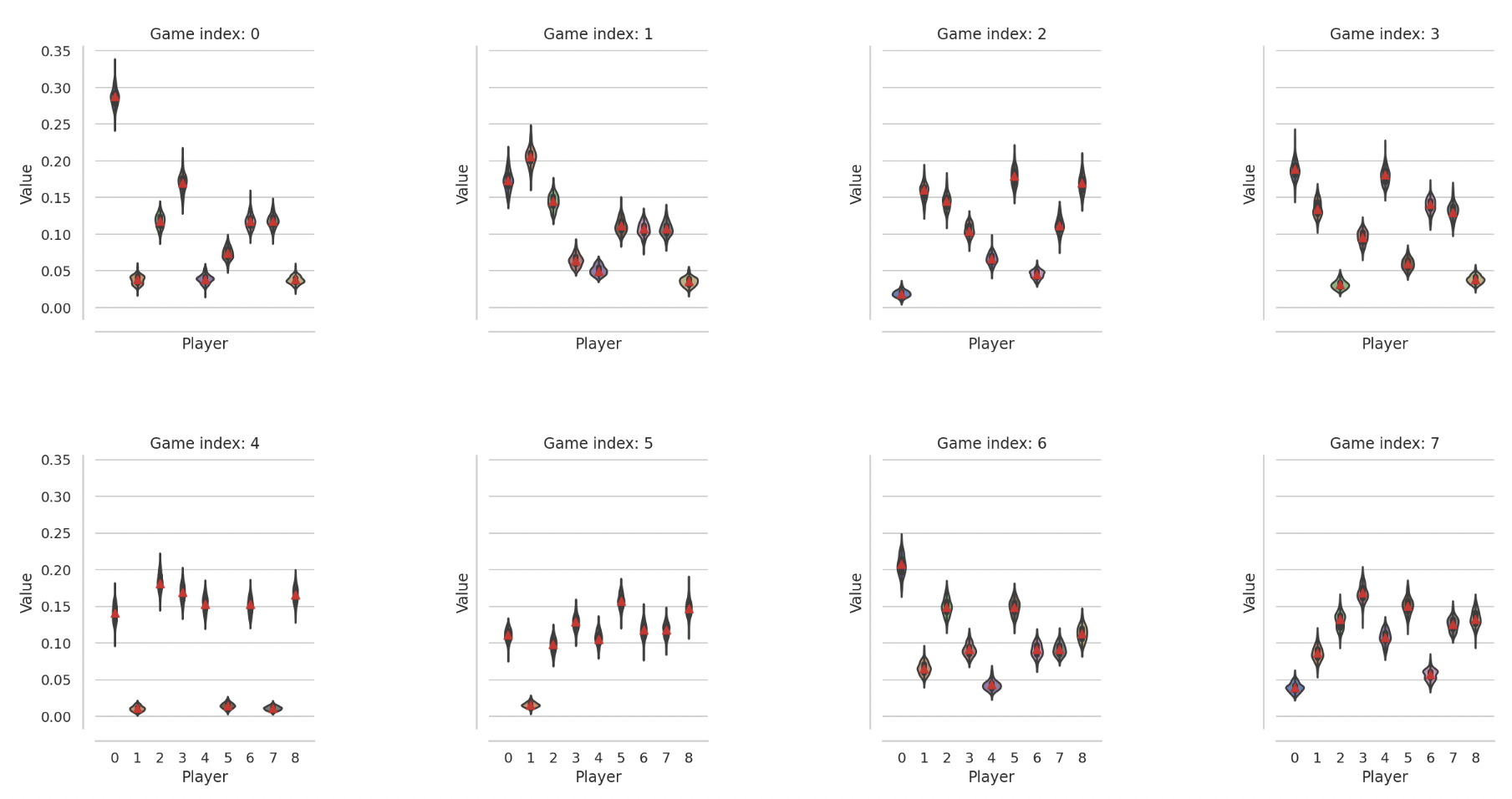}
     \end{subfigure}
     \hfill
     \begin{subfigure}[b]{0.48\textwidth}
         \centering
         \includegraphics[width=1\textwidth]{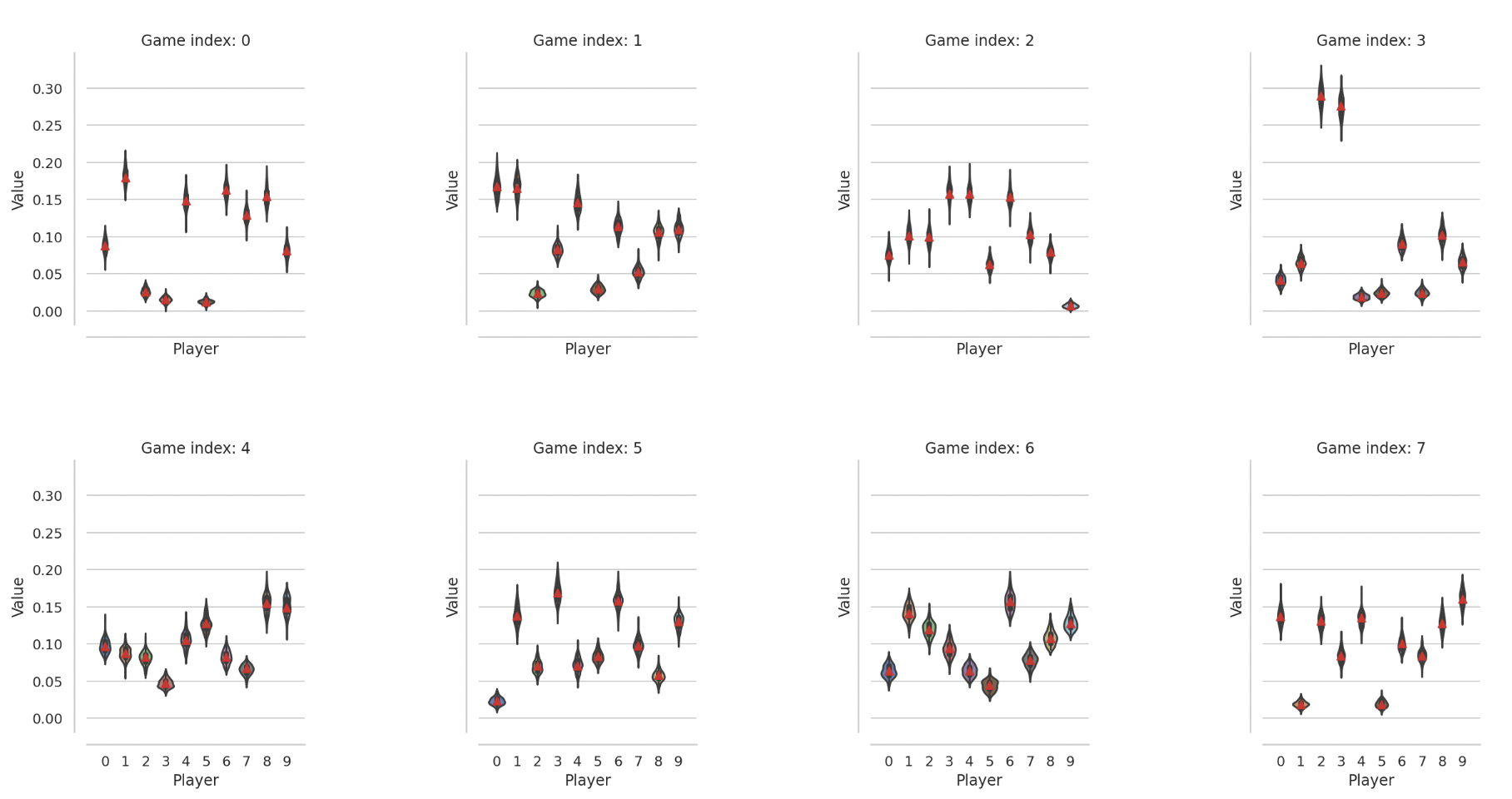}
     \end{subfigure}
        \caption{\textbf{Approximating ground-truth Shapley values.} Ten randomly sampled games with distribution for each player in the game, the red dot indicates the true Shapley value (i.e. when computing the marginal value for each permutation).}
        \label{fig:Shap_approx}
\end{figure}

\paragraph{Approximated ground-truth Shapley values.} \label{sec:approx_shap_values} To obtain the ground-truth solutions for Shapley values above nine players, we perform Monte-Carlo sampling to obtain a subset of all possible permutations. For each of the $R=10$ resamples we randomly sample $P=1000$ permutations and average across the resamples to obtain approximations of the ground-truth. To validate that the approximated labels are representative we compute the Mean MAE between the true and approximated Shapley values for $n=5$ up to $n=10$ players with 1000 games per $n$-player game. Across all $n$-player games the mean MAE between the true and approximated Shapley values was $0.0011$ and never exceeded $0.0014$. Figure \ref{fig:Shap_approx} shows the distributions per player together with the true label (red dot) for 8 and 9 player games.

\begin{figure}[h!t]
    \centering
    \includegraphics[width=0.6\textwidth]{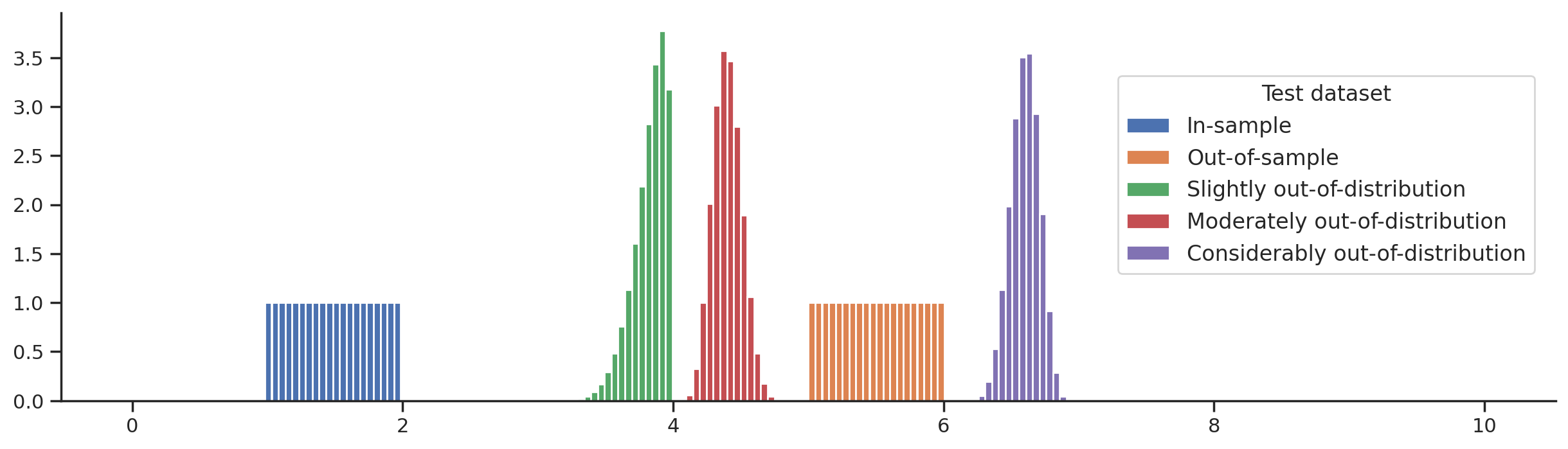}
    \caption{\textbf{Test data distributions in perspective.} To assess how robust models' are to shifts in the weight distribution, we re-parameterize the weight distribution in five  different ways.}
    \label{fig:test_distributions}
\end{figure}

\paragraph{Parameterizations for test datasets.} Figure \ref{fig:test_distributions} shows the five standardized test distributions that are generated with the parameters in Table~\ref{tab:parameter_testgames}.

\begin{table}[ht]
\centering
\caption{Test distributions are generated by the following parameterizations of the beta distribution}
\vspace{2mm}
\begin{tabular}{llll}
\toprule
Test set name & $\alpha$ & $\beta$ & location \\
\midrule
      In-sample               &    1    &  1    &   1      \\
      Out-of-sample               &    1    &  1    &   2.5N   \\
      Slightly out-of-distribution     &    8    &  12   &   2      \\
      Moderately out-of-distribution         &    7    &  1.5  &   1.5N   \\
      Considerably out-of-distribution               &    12   &  8    &   3N      \\  
\bottomrule
\end{tabular}
\label{tab:parameter_testgames}
\end{table}

\subsection{Explainable AI} \label{sec:explainable_ai}

We provide a step-wise explanation of how we performed the experiment in our explainable AI section, that is, train a neural network to predict the Shapley values for a dataset of features:
\begin{enumerate}
    \item \textbf{Select a dataset.} The first step is to select a dataset of choice. In our example, we end up with a feature matrix $\mathbf{X}^{N \times F}$ where each row is an observation and each column a feature, and a target vector $\mathbf{y} \in \mathbb{R}^N$ or  $y \in [0, 1]$ depending on the task.
    \item \textbf{Feature to target.} Following, we estimate a model $f_{\boldsymbol{\theta}}: \mathbb{R}^F \to \mathbb{R}$ that predicts one output for each set of features. 
    \item \textbf{Compute Shapley values through sampling.} We use the trained model from step 2 to obtain Shapley values with the SHAP package KernelExplainer\cite{lundberg2017unified}, which is based on LIME \cite{lundberg2017unified}. This yiels a matrix of Shapley values, one for each feature in every data instance: $\mathbf{\Phi} \in \mathbb{R}^{N \times F}$.
    \item \textbf{Feature to Shapley.} Finally, we create an array of $T=999$ linearly spaced train-test splits ${\mathbf{t} = (0.001, 0.002, \dots, 0.998, 0.999)}$. Each element $t_i$ in $\mathbf{t}$ indicates the ratio of samples that are used for training the neural network, with the remaining $(1 - t_i)N$ samples used for evaluation. For each train-test partition in $\mathbf{t}$ then, we train a feed forward neural network on $t_iN$ samples by minimizing the MSE between the predicted and actual Shapley values (full details in Table \ref{tab:XAI_model_params}) for 100 epochs, and evaluate on the $1-t_iN$ remainder (unseen) samples. 
\end{enumerate}

\paragraph{Datasets and preprocessing.} Table \ref{tab:XAI_datasets} summarizes the considered datasets. The UCI Banking dataset contains 15 features and a binary target (whether the client has subscribed a term deposit), whereas the Melbourne housing dataset contains 21 features and a regression target (the price of a house). We exclude the ``duration" feature in the Banking dataset to prevent data leakage, as is recommended in the documentation. Moreover, we exclude the ``Rooms'' feature as there is a lot of overlap in this and the ``Bedroom2'' feature. The numerical values in both datasets are z-scored, and categorical features are encoded as numbers. Missing values are encoded as zeros, so that the complete dataset can be used for our experiment.

\paragraph{Feature to target.} We partition the dataset into train and test and train a Random Forest Classification model on the UCI Banking set, obtaining a final test accuracy of 72 \%. Moreover, we train a decision tree regression model to predict the price given the set of house features and obtain a final MSE of 1.21 on the test dataset. 

\paragraph{Compute Shapley values.} We use the trained models to obtain Shapley values for each instance in the feature dataset $\mathbf{X}$ using the SHAP kernel explainer object \cite{lundberg2017unified}.

\paragraph{Feature to Shapley: procedure.} Finally, we can train a neural network using the datasets $\mathbf{X}, \mathbf{\Phi}$. To investigate the sample complexity required to learn a representative mapping from features to Shapley values, we conduct the following experiment. First, we create an array of $T=999$ linearly spaced training splits ${\mathbf{t} = (0.001, 0.002, \dots, 0.998, 0.999)}$. Each element in $t_i$ in $\mathbf{t}$ indicates the ratio of samples that are used for training the neural network, leaving the remaining $(1 - t_i)N$ samples for evaluation. For each train/test partition in $\mathbf{t}$ then, train a feed forward neural network on $t_iN$ samples through minimizing the MSE between the predicted and actual Shapley values (architecture and hyperparameters specified in Table \ref{tab:XAI_model_params}) for 100 epochs, and evaluate on the $1-t_iN$ samples. Figure \ref{fig:Explainable_ai_shap} depicts the resulting RMSE for both datasets as a function of the training fraction.


\begin{table}[ht]
\centering
\caption{Hyperparameters used for XAI experiments.}
\label{tab:XAI_model_params}
\vspace{2mm}
\resizebox{0.3\textwidth}{!}{%
\begin{tabular}{ll}
\toprule
\multicolumn{2}{l}{\textbf{Hyperparameter}} \\
\midrule
 Number of layers    &       3              \\
 Hidden size         &      128             \\
 Adam learning rate  &      1e-4            \\
 Adam $\epsilon$     &      1e-5            \\
 Dropout rate        &      0.1             \\
\bottomrule
\end{tabular}}
\end{table}

\begin{table}[ht]
\centering
\caption{Number of features pre and post processing for each considered dataset.}
\label{tab:XAI_datasets}
\resizebox{0.4\textwidth}{!}{%
\begin{tabular}{lllll}
\toprule
\textbf{Dataset} & \multicolumn{2}{l}{Number of features} & \multicolumn{2}{l}{Number of observations} \\
 & Pre & Post & Pre & Post \\
\midrule
 UCI Banking \cite{moro2014data}        & 15  &  14  &   11162  &  11162 \\
 Melbourne Housing \cite{pino2018melbourne} &  20   &  20   &  34857  &   34857   \\
\bottomrule
\end{tabular}%
}
\end{table}
\clearpage
\section{Additional Methods}

\begin{figure}[h!t]
    \centering
    \includegraphics[width=1\textwidth]{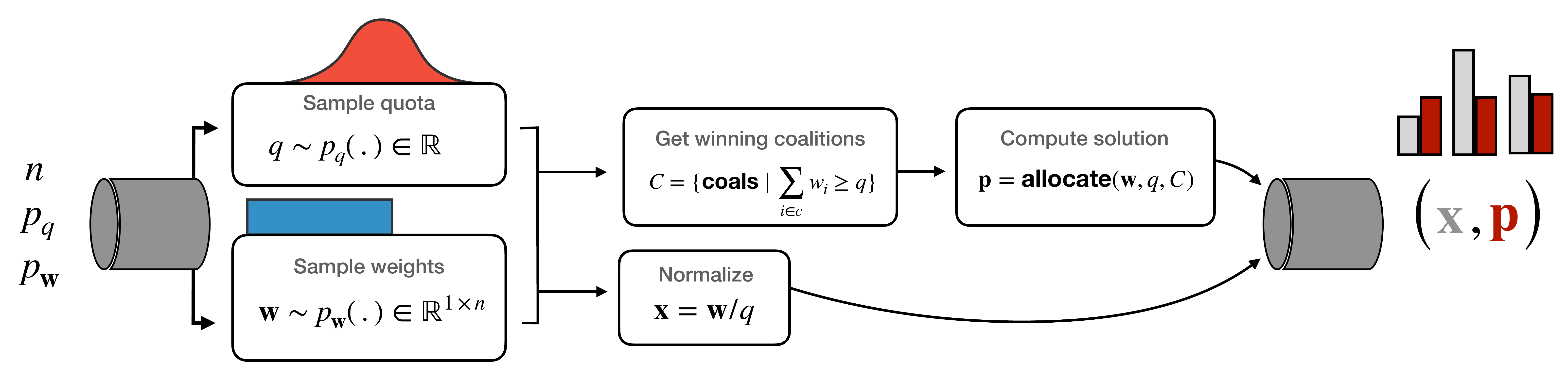}
    \caption{\textbf{Procedurally generating and solving one weighted voting game.} We sample the weights and quota to create a weighted voting game. The model inputs are divided by the quota and the payoffs are computed by solving the game, provided a solution concept. One game instance consists of a set of normalized weights $\mathbf{x}$ and the allocated payoffs $\mathbf{p}$.}
    \label{fig:data_gen}
\end{figure}

\section{Additional Results}
\paragraph{In-sample predictive performance.} For completeness, we provide the MAEs for each solution concepts on the in sample dataset up to $n=15$ players in Table \ref{tab:Predictive_performance_insample}.

\begin{table}[ht]
\centering
\caption{\label{tab:Predictive_performance_insample} Predictive performance across solution concepts on in-sample test dataset.}
\vspace{2mm}
\resizebox{0.6\textwidth}{!}{%
\begin{tabular}{lllllllll}
\toprule
& \multicolumn{2}{c}{\textbf{Least core payoffs}} & \multicolumn{2}{c}{\textbf{Least core excess}} & \multicolumn{2}{c}{\textbf{Shapley values}} & \multicolumn{2}{c}{\textbf{Banzhaf indices}} \\\midrule
Metric & \multicolumn{2}{c}{Mean MAE} & \multicolumn{2}{c}{MAE} & \multicolumn{2}{c}{Mean MAE} & \multicolumn{2}{c}{Mean MAE} \\
N & Fixed & Variable & Fixed & Variable & Fixed & Variable & Fixed & Variable \\ \midrule
4     &  0.087 &  0.140 &   0.034 &  0.058 &  0.069 &  0.072 &  0.067 & 0.097  \\
5     &  0.077 &  0.096 &  0.031 &   0.045 &  0.040 &  0.047 &  0.035 & 0.060  \\
6     &  0.058 &  0.069 &   0.028 &  0.033 &  0.026 &  0.031 &  0.021 & 0.037  \\
7     &  0.040 &  0.051 &  0.021 &   0.027 &  0.022 &  0.023 &  0.014 & 0.024 \\
8     &  0.027 &  0.036 &  0.015 &   0.021 &  0.016	&  0.018 &  0.015 & 0.017  \\
9     &  0.019 &  0.031 &  0.012 &   0.019 &  0.010	&  0.015 &  0.012 & 0.014  \\
10    & 0.014 &  0.023 &   0.007 &   0.017 &  0.009	&  0.013 &  0.006 & 0.013   \\
11    & 0.010 &  0.021 &   0.006 &   0.020 &  0.006	&  0.012 &  0.009 & 0.013   \\
12    & 0.009 &  0.018 &   0.005 &   0.023 &  0.005 &  0.011 &  0.004 & 0.013   \\
13    & 0.006 &   0.016 &   0.003 &  0.027 &  0.009	&  0.010 &  0.004 & 0.014  \\
14    & 0.005 &  0.014 &   0.003 &   0.032 &  0.004	&  0.009 &  0.009 & 0.014   \\
15    & 0.005 &  0.013 &   0.003 &   0.036 &  0.004	&  0.009 &  0.012 & 0.014  \\
\midrule
Avg & 0.019 & 0.015  &  0.029 & 0.051   &  0.018	&  0.022 & 0.017 & 0.027 \\
 \bottomrule
\end{tabular}%
}
\end{table}

\paragraph{Capturing step-wise jumps: full description.}\label{sec:stepwise_desc}
We provide a more detailed description to Figure \ref{fig:perturb_quota}. In the Figure on the \textbf{left}, each player is a winning coalition by itself when $q=1$. Up to $q=3$, player 3 is more critical than players 1 and 2 and receives a larger share of the joint payoffs. All players obtain equal payoffs when the only winning coalition is the grand coalition. On the \textbf{right} side: each player is a winning coalition by itself when $q=1$, which results in an equal payoff of $0.2$ for each player. Increasing the quota continuously alters the set of winning coalitions, thereby the relative importance of each player. At $q=3$ players 1, 2 and 3 have become less critical and obtain a payoff of $~0.18$ each compared to $~0.22$ for player 4 and 5. Finally, when $q = \sum_{i=1}^N w_i = 7$, the only winning coalition that can be formed is the grand coalition: each player is required and the joint payoff is divided equally.

\section{Hardware details}\label{sec:hardware}
We trained our neural networks on an internal compute cluster with 10 Quadro RTX 6000 GPUs and two \href{https://www.intel.com/content/www/us/en/products/sku/192444/intel-xeon-gold-5218-processor-22m-cache-2-30-ghz/specifications.html}{Intel(R) Xeon(R) Gold 5218 CPUs}, each of which has 16 physical cores (32 logical processors).

\section{Asset licensing}\label{sec:asset_licenses}

Two main assets were used in this work:
\begin{itemize}
    \item The UCI Banking dataset
    \item The Melbourne Housing dataset
\end{itemize}

The Melbourne Housing dataset was released under the license CC BY-NC-SA 4.0. The UCI Banking dataset is licensed under a Creative Commons Attribution 4.0 International (CC BY 4.0) license.

\section{Evaluating the Speedup Achieved by Our Approach, Depending on the Size of the Dataset}\label{sec:in_depth_shap_exp}

Many domains involve datasets with a large number of features, sometimes with over thousands or millions of data instances. A prominent example is language modelling, where large bodies of text are generated, translated or processed in  different manners \cite{patel2021game}. In such cases, computing the Shapley value for each data instance using sampling-based methods takes an extremely long time.

\begin{figure}[h!t]
     \centering
     \begin{subfigure}[b]{0.49\textwidth}
         \centering
         \includegraphics[width=1\textwidth]{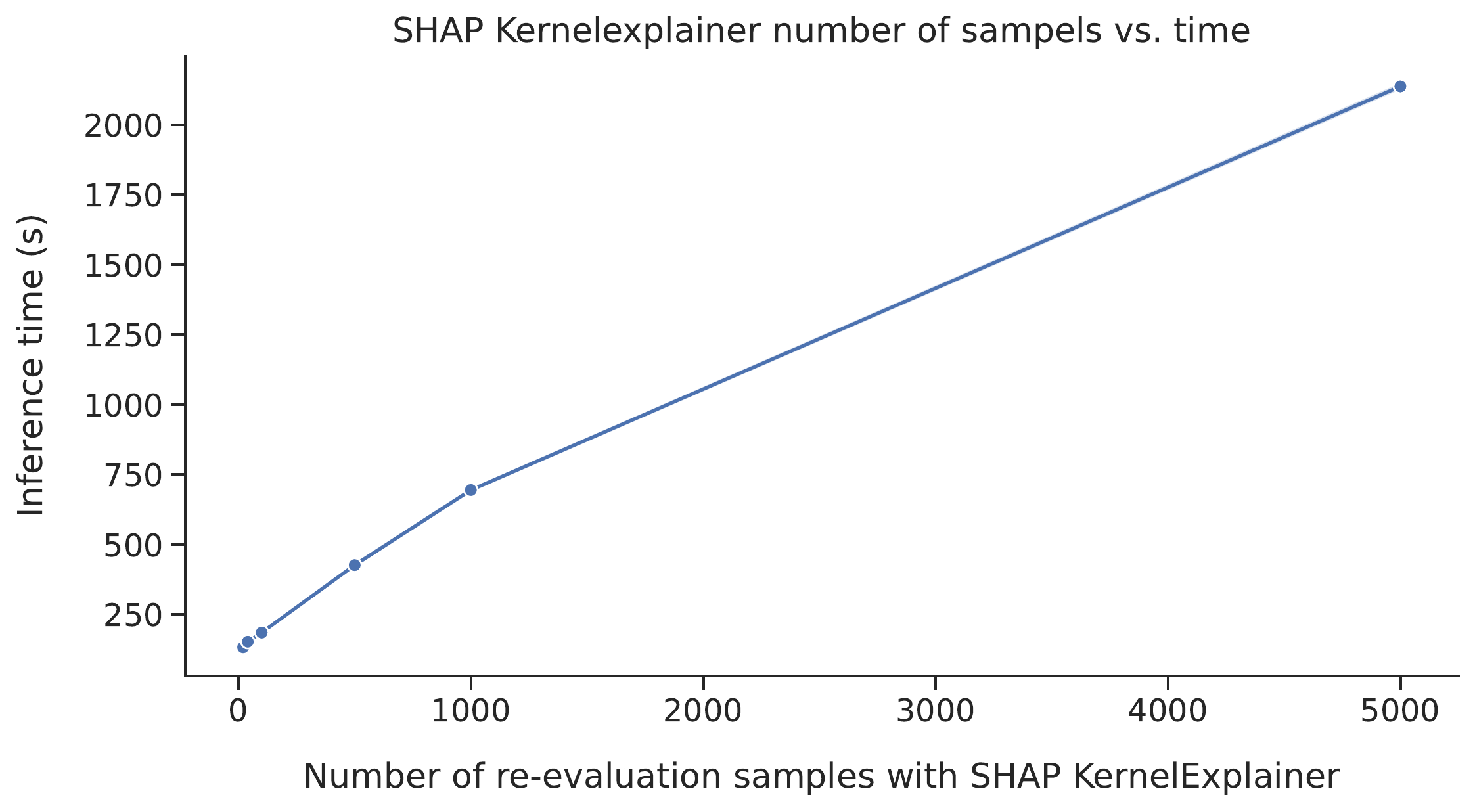}
     \end{subfigure}
     \hfill
     \begin{subfigure}[b]{0.49\textwidth}
         \centering
         \includegraphics[width=1\textwidth]{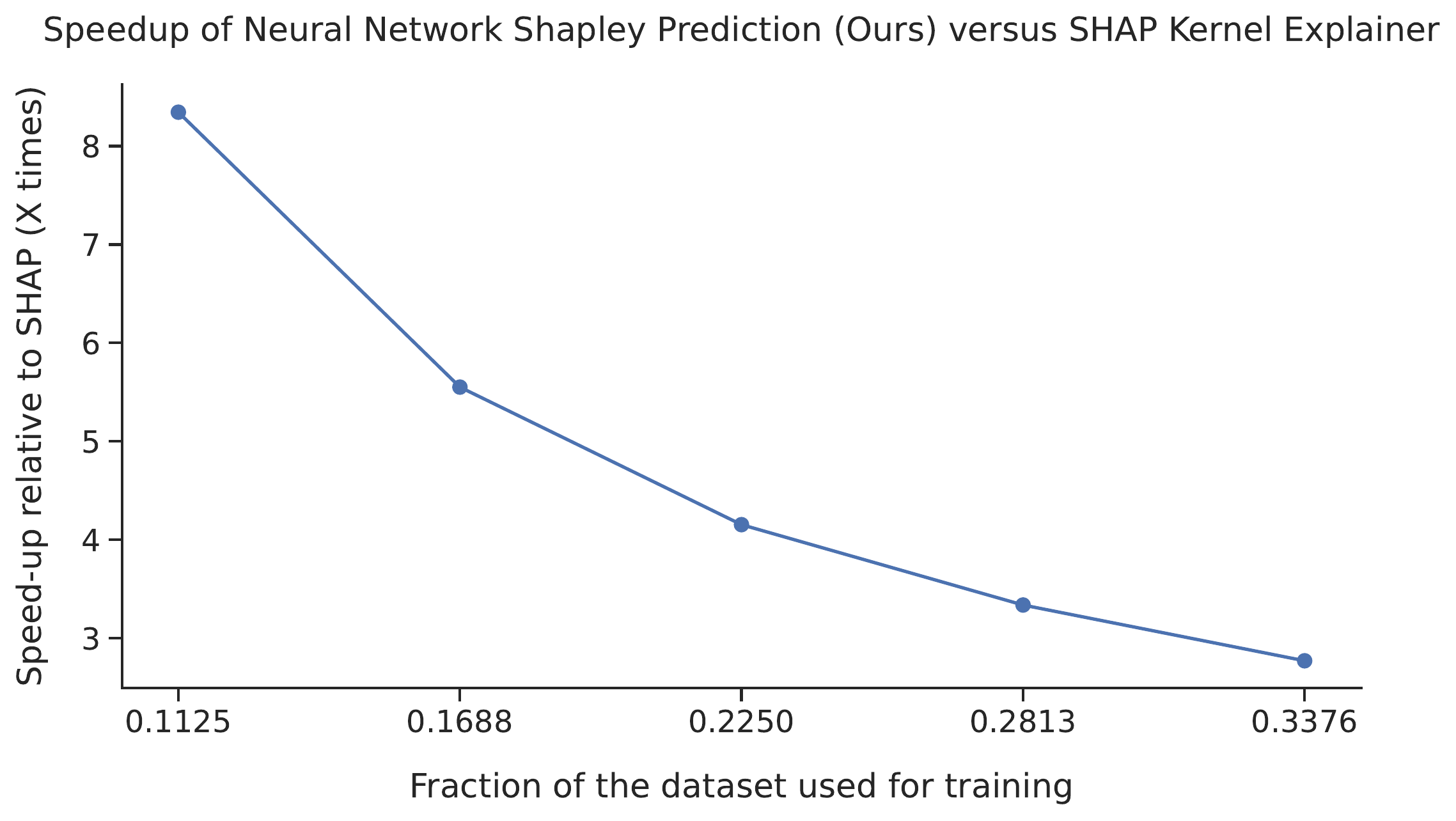}
     \end{subfigure}
        \caption{\textbf{Speeding up Shapley value computations.} \textit{Left}: The inference time, here the time it takes to compute the shapley values of 9000 instances, scales approximately linearly with the number of samples. \textit{Right}: The x times speedup provided by our model-based approach contrasted by the fraction of data used for training.}
         \label{fig:shap-exp}
\end{figure}

Our approach is targeted at speeding up feature importance computations of many instances. We perform a simple experiment to demonstrate the effectiveness of a model-based approach for explaining model decisions. We take the Melbourne Housing dataset and obtain $~$9000 instances by 13 features after preprocessing steps (encoding the categorical features and standardization). First, we generate a ground-truth dataset by setting the number of samples in the SHAP package (KernelExplainer) to 5000 samples. The number of samples in the KernelExplainer determines the number of times a model is re-evaluated for every prediction. A higher number of samples will result in more accurate (and lower variance in the) Shapley values, but require a longer compute time. Computing the Shapley values on our $~$9000 samples takes 4.13 hours. Figure \ref{fig:shap-exp} (left) shows that the inference time scales approximately linearly with the number of re-evaluation samples.

To contrast the difference in computation time, we compute Shapley values on 10 \% of the data (computation time is 29 minutes) train the neural network in 30 seconds, and predict Shapley values on the remaining 90\% of the data (less than a second). All together, this procedure takes roughly 12 \% of the time of it took to compute Shapley values for the entire dataset. This speeds-up the whole procedure by 8x while keeping a reasonable prediction quality on the remaining unseen data (Figure \ref{fig:shap-exp} right). 

Next, we use SHAP again to compute Shapley values for 90\% of the data, ~$8100$ instances, this time with the default number of model re-evaluations samples, yielding slightly accurate Shapley values (the default is $2 \times \text{number of features} + 2048$). We quantify the trade-off in prediction quality as the by comparing the model predicted Shapley values to the Shapley values obtained with the default number of samples.

Concretely, we measure the Mean Mean Absolute Error (MMSE) of our model predictions and the SHAP package to the ground-truth dataset. We note that speedup does come at a slightly reduced empirical performance: the model MSE on the test set is, $\text{MSE} = 1.15 \times 10^{-3}$, compared to the MSE with SHAP: $\text{MSE} = 7.4 \times 10^{-5}$. However, we expect that hyperparameter tuning and better preprocessing will improve the quality of the model predictions.

All together, it is easy to see how this win in computation time translates to larger datasets. Consider a similar dataset with 100,000 or a million entries. Computing Shapley values on the entire dataset with SHAP will take weeks (40 or 400 hours), while the time for our procedure remains relatively the same: building the train set and training the neural network  only takes a fraction of the time required by the sampling-based approach: the analysis using our method would take a few hours. 

\section{Heuristics and Simpler Models}

We now consider a common heuristic for payoff allocation in weighted voting games, the {\it weight proportional} payoff allocation. We use this as a benchmark to evaluate our models' performance against. We find that our fixed models outperform this benchmark heuristic across solution concepts and test sets. 


\paragraph{WeightProportional baseline.}\label{eq:heuristic} The WeightProportional (WP) heuristic divides the payoffs in a way that is directly proportional to the players' weights, that is, the payoff of a player $j$ is:
\begin{equation}
    \mathbf{p}_{j} = \frac{w_j}{\sum_{i=1}^n w_i} 
\end{equation}
for any $n-$player game. WP provides meaningful benchmark as it is the most naive method for allocating payoffs. The resulting payoffs (and thus errors) do not take into account the quota, which may have a major impact on the correct solutions by means of the emergent set of winning coalitions. As such, any improvement on this benchmark thus suggests a better understanding of the collaborative structures in the game at hand.

We also compare the predictive performance of our models against the predictive performance of multinomial regression models. These models have just one layer with $n$ inputs and $n$ outputs for each $n-$player game and a softmax nonlinearity at the end. Across all game sizes, our models outperform the linear models by 45 \% to 95 \% percent. Results are found in Table \ref{tab:linear-model-comp}.

\renewcommand{\arraystretch}{1.3}
\begin{table}[h!t]
\centering
\caption{\label{tab:linear-model-comp} Comparing neural network performance to multinomial logistic regression models.}
\vspace{2mm}
\resizebox{1\textwidth}{!}{%
\begin{tabular}{lllllll}
\hline
 & \textbf{Linear Core} & Neural net Core & \textbf{Linear Shapley} & Neural net Shapley & \textbf{Linear Banzhaf} & Neural net Banzhaf \\ \hline
\multicolumn{1}{l|}{N}  & \multicolumn{6}{l}{}                           \\ \hline
\multicolumn{1}{l|}{4}  & 0.212 & 0.089 & 0.137 & 0.07  & 0.116 & 0.069  \\
\multicolumn{1}{l|}{5}  & 0.147 & 0.078 & 0.098 & 0.041 & 0.083 & 0.035  \\
\multicolumn{1}{l|}{6}  & 0.124 & 0.059 & 0.080 & 0.026 & 0.071 & 0.021  \\
\multicolumn{1}{l|}{7}  & 0.108 & 0.041 & 0.070 & 0.022 & 0.063 & 0.014  \\
\multicolumn{1}{l|}{8}  & 0.102 & 0.028 & 0.065 & 0.015 & 0.059 & 0.014  \\
\multicolumn{1}{l|}{9}  & 0.089 & 0.020 & 0.056 & 0.009 & 0.051 & 0.012  \\
\multicolumn{1}{l|}{10} & 0.078 & 0.014 & 0.050 & 0.008 & 0.045 & 0.006  \\
\multicolumn{1}{l|}{11} & 0.072 & 0.010 & 0.045 & 0.006 & 0.042 & 0.009  \\
\multicolumn{1}{l|}{12} & 0.065 & 0.009 & 0.041 & 0.005 & 0.041 & 0.004  \\
\multicolumn{1}{l|}{13} & 0.062 & 0.006 & 0.036 & 0.009 & 0.035 & 0.004  \\
\multicolumn{1}{l|}{14} & 0.056 & 0.005 & 0.036 & 0.004 & 0.034 & 0.009  \\
\multicolumn{1}{l|}{15} & 0.057 & 0.005 & 0.032 & 0.004 & 0.032 & 0.0012 \\
\multicolumn{1}{l|}{16} & 0.051 & 0.003 & 0.029 & 0.004 & 0.030 & 0.003  \\
\multicolumn{1}{l|}{17} & 0.051 & 0.003 & 0.028 & 0.004 & 0.029 & 0.006  \\
\multicolumn{1}{l|}{18} & 0.047 & 0.005 & 0.028 & 0.007 & 0.027 & 0.011  \\
\multicolumn{1}{l|}{19} & 0.045 & 0.002 & 0.026 & 0.007 & 0.025 & 0.003  \\
\multicolumn{1}{l|}{20} & 0.044 & 0.003 & 0.025 & 0.006 & 0.024 & 0.006  \\ \hline
\end{tabular}%
}
\end{table}

\section{Hard cases}

Across solution concepts (Shapley, Banzhaf and Least core), discontinuous jumps in the solution space yield large errors (see Section \ref{sec:capture_disc} and Figures \ref{fig:perturb_quota}, \ref{fig:var-eu-application}). In this section, we highlight failure cases for the Least core. In contrast to the power indices, a solution in the least core is obtained by solving a Linear Program under several hard constraints. Thus, a solution to a weighted voting game is only feasible when, for each winning coalition $C \in C^{\text{win}}$, we have that
\begin{equation}\label{eq:feasibility}
    v(C) = \sum_{i=1}^C p_i \geq 1 - \varepsilon
\end{equation}
where $p_i$ is the payoff of player $i$ in coalition $C$. In words, each winning coalition must obtain joint reward that is at least as good as what they would have obtained on their own. If this criteria is not met, the solution is infeasible: some players will have an incentive to deviate from the grand coalition.

\begin{figure}[h!t]
    \centering
    \includegraphics[width=0.8\textwidth]{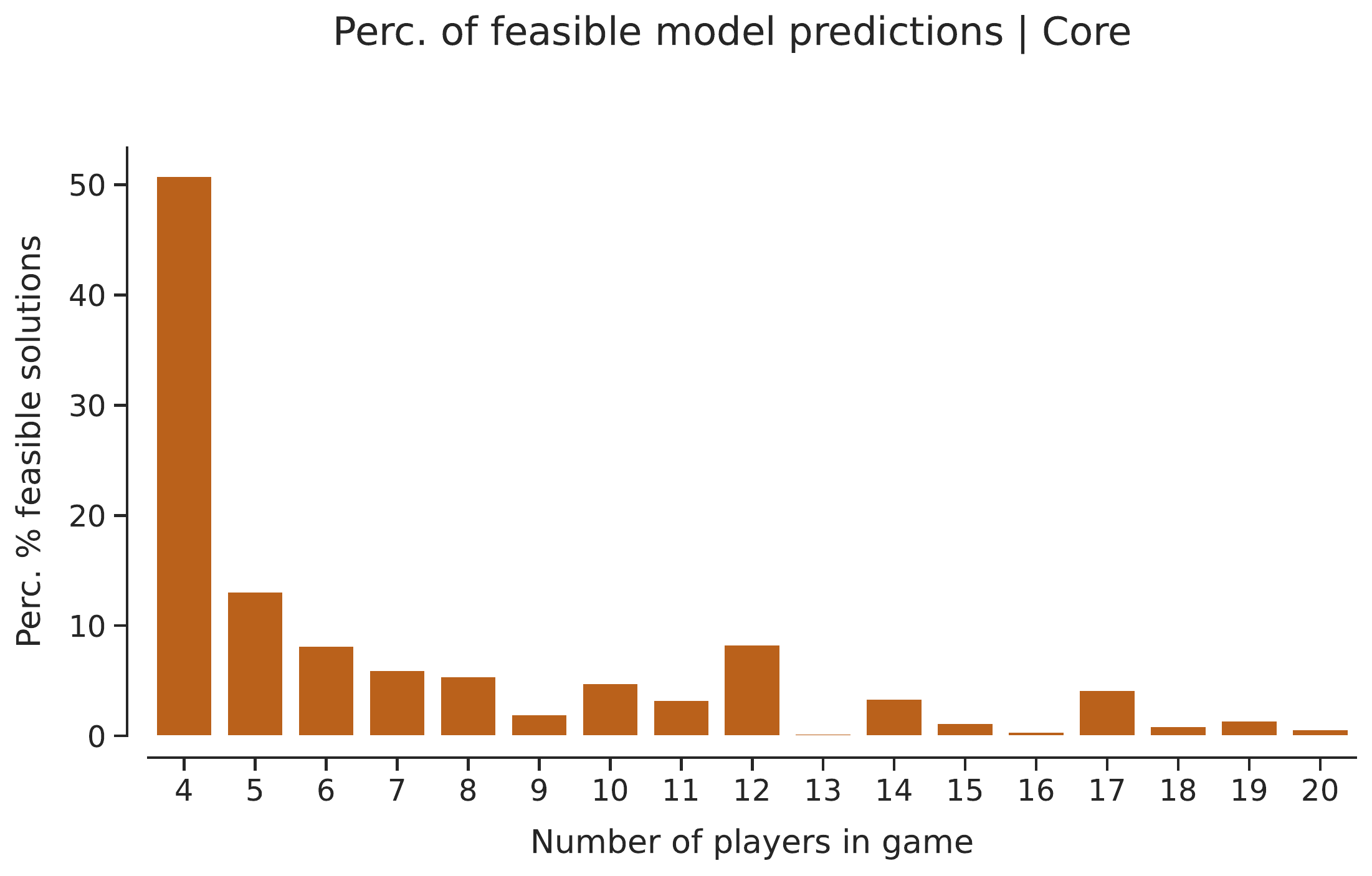}
    \caption{\textbf{Percentage of feasible solutions for each $n-$player game}. We count the number of solutions that meet the hard constraints by validating equation \ref{eq:feasibility} for each winning coalition $C$ in the set of winning coalitions across 1000 samples per $n-$player game.}
    \label{fig:perc-feasible}
\end{figure}

\textbf{Majority of the model predictions fail to meet hard constraints}. Figure \ref{fig:perc-feasible} depicts the percentage of feasible solutions over the game sizes. For small player games ($n < 8$), between 50 and 5.5 \% of the model predictions are feasible). The larger the number of players, the fewer model predictions satisfy the hard constraints. From eight player games and above, the number of feasible predictions ranges between 0.2 and 10 \%. A preliminary analysis shows no relationship between infeasible solutions and larger errors in the payoffs nor the LCV.

The fact that the model is not able to predict many feasible solutions in these large games is unsurprising. The set of winning coalitions increases approximately exponentially in $n$. As such, the more players in a game, the more hard constraints need to be satisfied in order for a solution to be feasible per game. All hard constraints need to be met in order for a solution to be feasible. We expect that including a penalty term for infeasible predictions during training will help to improve this metric and hope to address this in future work.

\begin{figure}[h!t]
    \centering
    \includegraphics[width=1\textwidth]{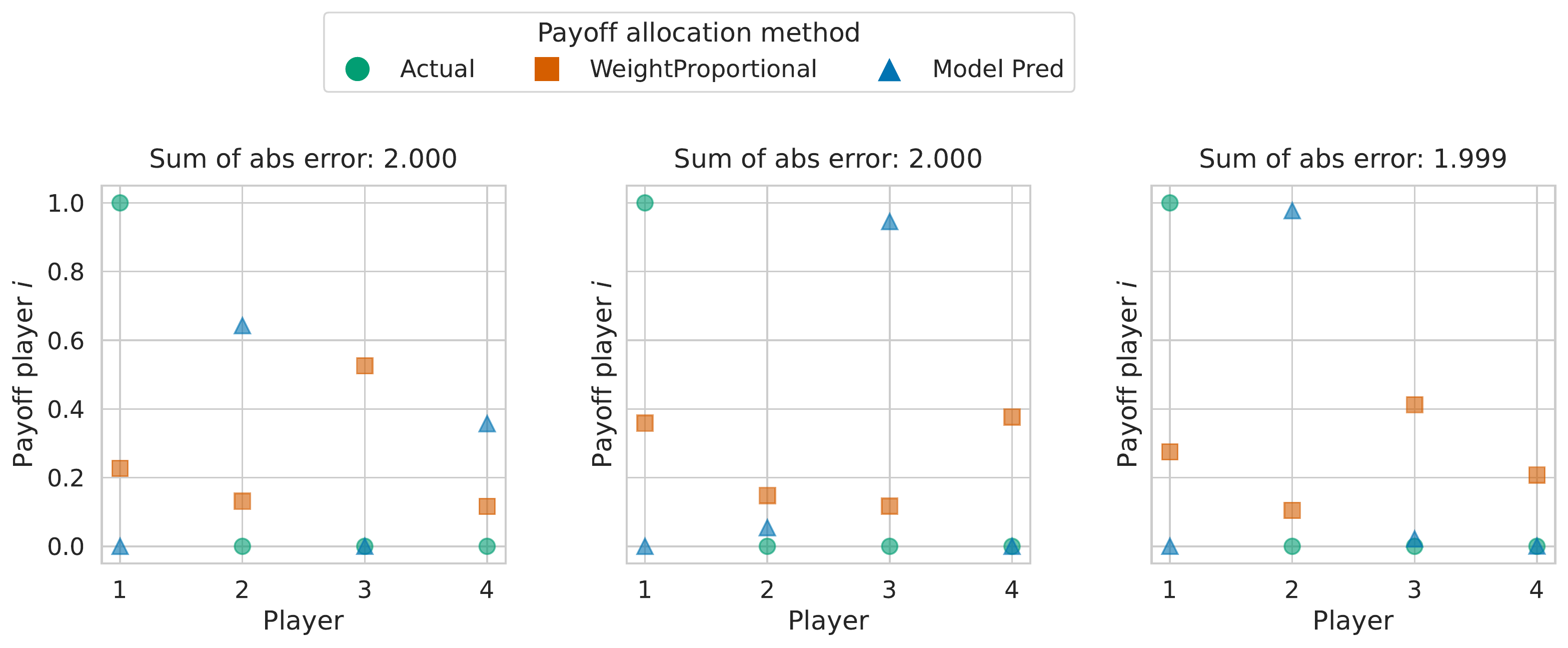}
    \caption{\textbf{Examples of games where the joint payoff is allocated to a single player}. We depict the actual solutions (green dots), payoffs allocated according to the weight proportional heuristic (see eq. \ref{eq:heuristic}, orange squares) and model predictions (blue triangles).}
    \label{fig:example-corner-solutions-core}
\end{figure}

\begin{figure}[h!t]
    \centering
    \includegraphics[width=0.55\textwidth]{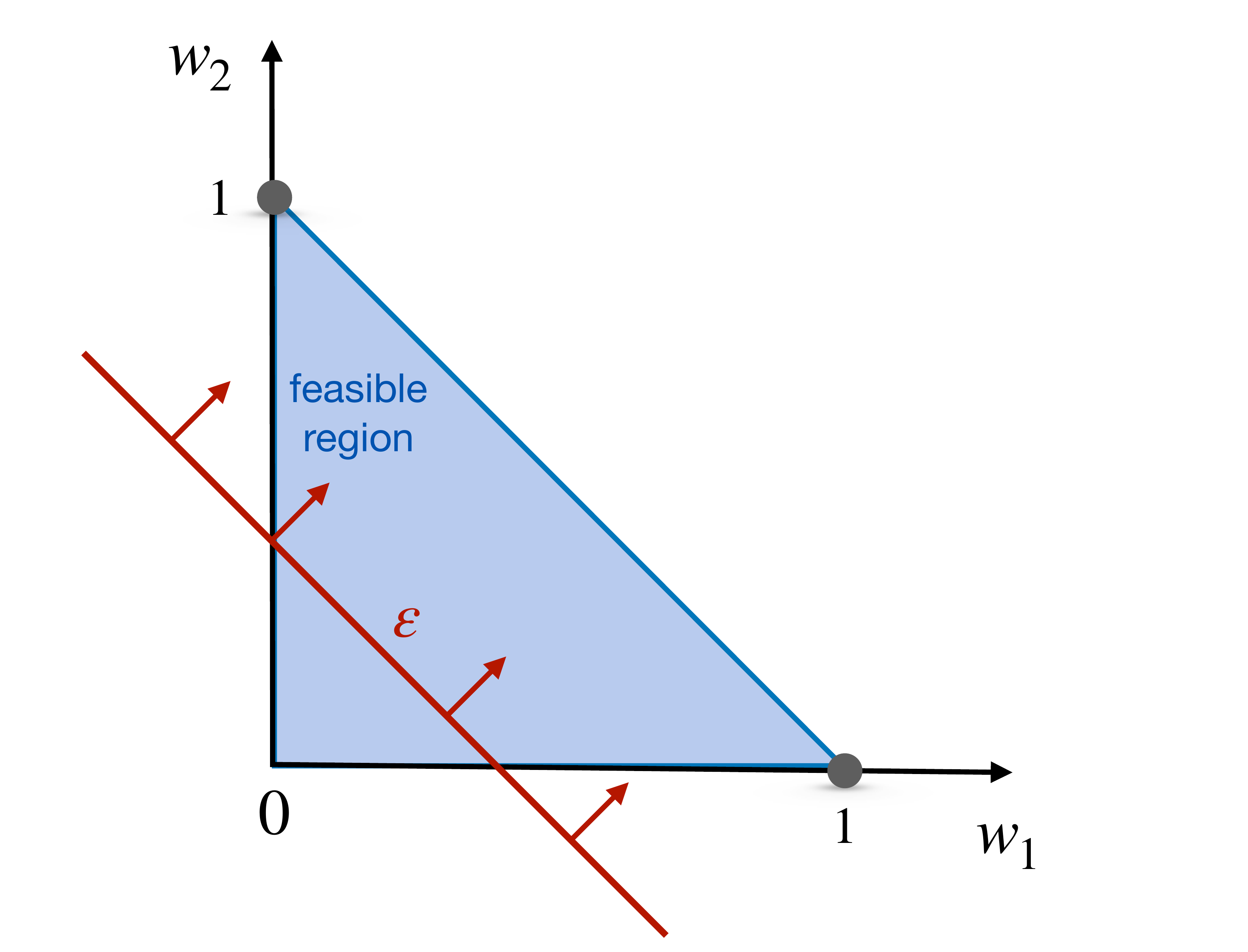}
    \caption{\textbf{Visual depiction of the corner solutions and feasible region of a two-player weighted voting game}. Take the weighted voting game $\mathbf{w} = (1, 1)$ with $q = 2$. Simplex based methods tend to return corner solutions, in this case $\mathbf{p} = (1, 0)$ or $\mathbf{p} = (0, 1)$ with a LCV $=0$. }
    \label{fig:corner-solutions-core}
\end{figure}

\paragraph{When the least core is a set, simplex based methods tend to find corner solutions, which leads to ambiguity in the solutions.} Further analysis of individual weighted voting games reveals that the most problematic games are cases where the least core contains multiple correct solutions, but the full joint payoff is allocated to a single player. In other words, payoffs are divided such that one player receives one, and all the others zero. To illustrate what is happening here, consider a simple four player game that produced the largest error (Figure \ref{fig:example-corner-solutions-core}). We focus our analysis on the game on the left, but our insights apply to a wide range of $n$-player games in the dataset. This particular game has the following weights and quota (rounded to one decimal place for simplicity)
\begin{equation}
    \mathbf{w} = (2.8, 1.6, 6.6, 1.5) ; \quad q = 12.1
\end{equation}
where we can see that no player is a winning coalition by itself (all $w_i$'s $<$ q). In fact, the only winning coalition in this game is the grand coalition, that is, 
\begin{equation}
    C^{\text{min}} = \{ 1, 2, 3, 4 \}
\end{equation}
The solution to this weighted voting game is:
\begin{align}
    \mathbf{p} = (1, 0, 0, 0) \quad \text{with} \quad \text{LCV} = 0 
\end{align}
where player one receives the full joint payoff, even while its' weight is smaller than the weight of player three ($2.8 < 6.6$). The reason why we obtain this solution comes down to two facts. First of all, the least core is a set, and can therefore contain multiple correct solutions. Secondly, simplex based methods return corner solutions. Thus, in cases where several players are required to form a winning coalition, the solver will allocate the joint payoff to an arbitrary player that belongs to this critical subset. 
Figure \ref{fig:corner-solutions-core} depicts this visually for a simple two player game. Here, we have the WVG $\mathbf{w} = (1, 1)$ with $q=2$, so the only winning coalition is the grand coalition. Solving this game results in either of the corner solutions: the payoff vector $\mathbf{p} = (1, 0)$ or $\mathbf{p} = (0, 1)$ with a LCV $=0$. However, if we define $\alpha$ as the solution of a player one, any combination $\mathbf{p} = (\alpha, 1-\alpha)$ is a feasible solution in the least core. 

Obtaining the full set of solutions for this game can be done empirically. In principle, we can take the payoff vector and LCV above, generate the set of permutations of length two (player pairs) and check whether the solution is still feasible if we give a small value $\delta$ from one player (whose payoff is not zero) to any other player. However, this approach does not scale to larger games. It would be more efficient to derive rules for acquiring the full least core set from a single solution.

We limited our analysis above to four-player games. However, we observe the above behavior irrespective of the number of players in the game. Therefore, addressing this issue is critical for improving the predictive performance of our models.

\section{Applying Our Models to Voting In The European Council}
In this section we apply our trained models to predict voting power of member states in the EU Council. First, we take the weights from four countries in the EU voting council \cite{leech2002designing}, see Table \ref{tab:EU-council-weights}, and define the quota as the majority vote (50 \% of the total weights + 1). Our (weighted) voting game consists of the member states Hungary, Netherlands, Poland and Ireland with the weights $\mathbf{w} = (12, 13, 27, 7)$, and quota $q = 30,5$. 

\begin{table}[h!t]
\centering
\caption{\label{tab:EU-council-weights} Comparison of voting weights in the Council of the European Union \cite{leech2002designing}.}
\vspace{2mm}
\resizebox{0.85\textwidth}{!}{%
\begin{tabular}{lllrl}
\toprule
Member state & Population & Population Perc. &  Weights  & Weights Perc. \\
\midrule
     Germany &     82.54m &            16.5\% &       29 &          8.4\% \\
      France &     59.64m &            12.9\% &       29 &          8.4\% \\
          UK &     59.33m &            12.4\% &       29 &          8.4\% \\
       Italy &     57.32m &            12.0\% &       29 &          8.4\% \\
       Spain &     41.55m &             9.0\% &       27 &          7.8\% \\
      Poland &     38.22m &             7.6\% &       27 &          7.8\% \\
     Romania &     21.77m &             4.3\% &       14 &          4.1\% \\
 Netherlands &     17,02m &             3.3\% &       13 &          3.8\% \\
      Greece &     11.01m &             2.2\% &       12 &          3.5\% \\
    Portugal &     10.41m &             2.1\% &       12 &          3.5\% \\
     Belgium &     10.36m &             2.1\% &       12 &          3.5\% \\
  Czech Rep. &     10.20m &             2.1\% &       12 &          3.5\% \\
     Hungary &     10.14m &             2.0\% &       12 &          3.5\% \\
      Sweden &      8.94m &             1.9\% &       10 &          2.9\% \\
     Austria &      8.08m &             1.7\% &       10 &          2.9\% \\
    Bulgaria &      7.85m &             1.5\% &       10 &          2.9\% \\
     Denmark &      5.38m &             1.1\% &        7 &          2.0\% \\
    Slovakia &      5.38m &             1.1\% &        7 &          2.0\% \\
     Finland &      5.21m &             1.1\% &        7 &          2.0\% \\
     Ireland &      3.96m &             0.9\% &        7 &          2.0\% \\
\bottomrule
\end{tabular}
}
\end{table}

\begin{figure}[h!t]
    \centering
    \includegraphics[width=0.8\textwidth]{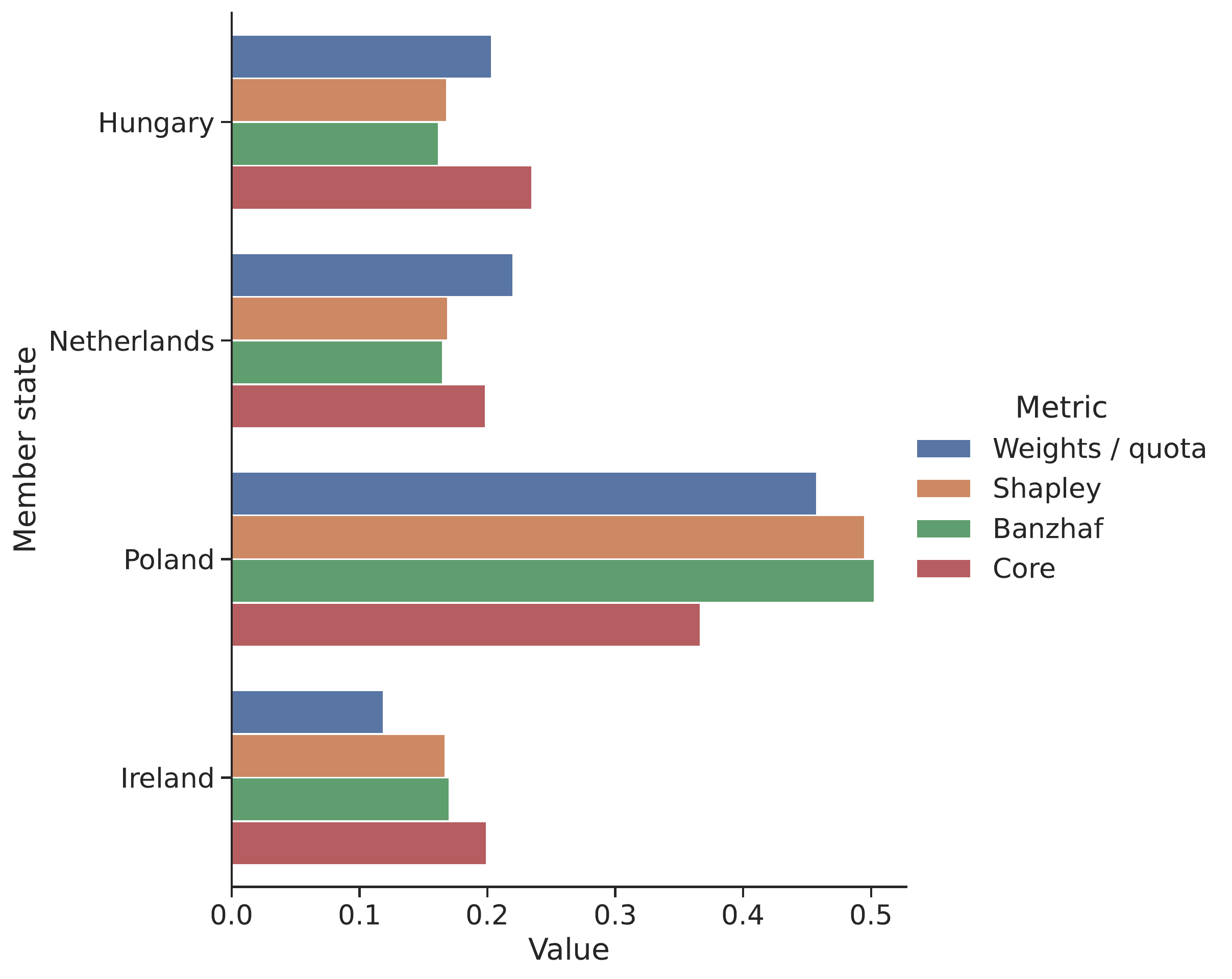}
    \caption{\textbf{Allocating payoffs to member states in the EU Council.} A comparison of how absolute power (the normalized weights ($\mathbf{w} / q$)) relates to the allocated share of the joint payoff.}
    \label{fig:voting-application}
\end{figure}

Following, we divide the weights by the quota and feed the normalized weights (see Section \ref{modelanddata}). We contrast the model predictions against the ground-truth solutions and find that the Mean Mean Absolute Error (MMAE) between our model predictions and actual solutions per member state is 0.003 (Banzhaf), 0.002 (Shapley), 0.017 and 0.007 for the Least core payoffs and excess respectively. Figure \ref{fig:voting-application} compared the payoff allocations for the Shapley, Banzhaf and Core solution concepts. We observe that the contribution measured by the Power indices (Shapley and Banzhaf) is less than the weight, while the Core attributes a larger share of the joint payoff to this member state.

\begin{figure}[h!t]
    \centering
    \includegraphics[width=1\textwidth]{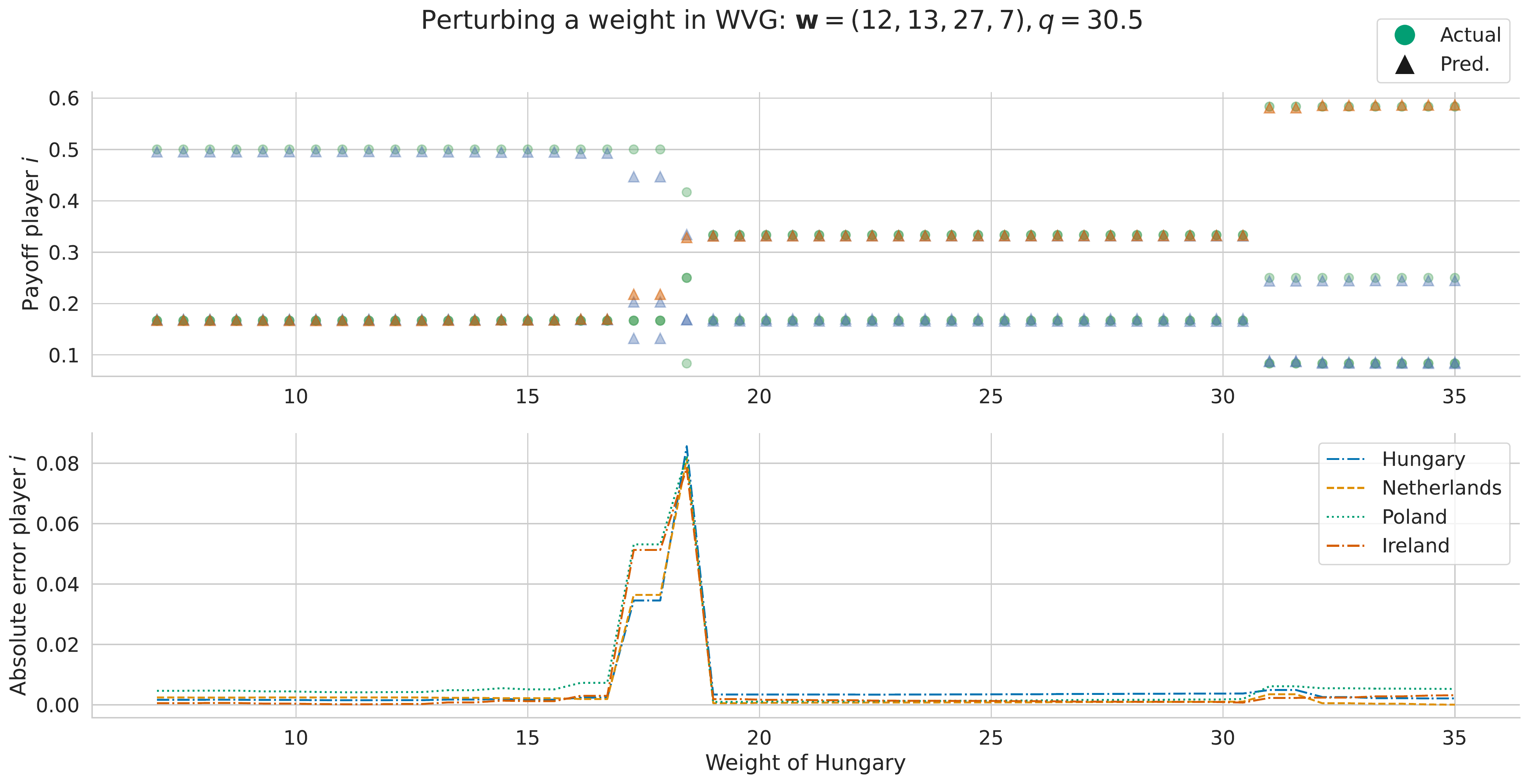}
    \caption{\textbf{Perturbing the weight of a member state in a voting game.} \textit{Top}: Model predictions and actual payoffs for a four-player voting game with weight vector $\mathbf{w} = (12, 13, 27, 7)$. textit{Bottom}: Absolute errors between the individual member state payoffs.}
    \label{fig:weight-perturbation}
\end{figure}

Next, we perform a sensitivity analysis. In such an analysis, we are interested in the interdependence of member states' voting power. Holding all other game parameters  constant, we increment the weight of Hungary ($w_\text{Hungary} = 7$) with values of one, until it exceeds the quota of 30,5. Figure \ref{fig:weight-perturbation} depicts how the member states' voting power (in Shapley value) changes as Hungary gains absolute power (as we increase the weight). The predicted payoff for Hungary is depicted by the red triangles, the model predictions for the other states in blue, and the actual solutions by the green dots. We observe that there are two transition points in the solution space. One transition point occurs when $w_\text{Hungary} = 17.5/18.5$. Hungary can now form a winning coalition both with the Netherlands ($w_\text{Netherlands} = 12$) and with Poland ($w_{\text{Poland}} = 13$), which raises its power in the voting and accordingly its' share of the joint payoff.  We notice that our model is able to predict accurate payoffs overall, with a spike in absolute error per player at this transition point.

The second transition point occurs when $w_\text{Hungary} = 30.5 = q$. Hungary is now also a winning coalition by itself, thereby obtaining a larger share of the joint payoff. We observe that its' Shapley value increases to approximately 0.6, obtaining the more than half of the total reward. Our model is able to capture this transition almost perfectly (MMAE $< 0.01$).

\begin{figure}[h!t]
    \centering
    \includegraphics[width=1\textwidth]{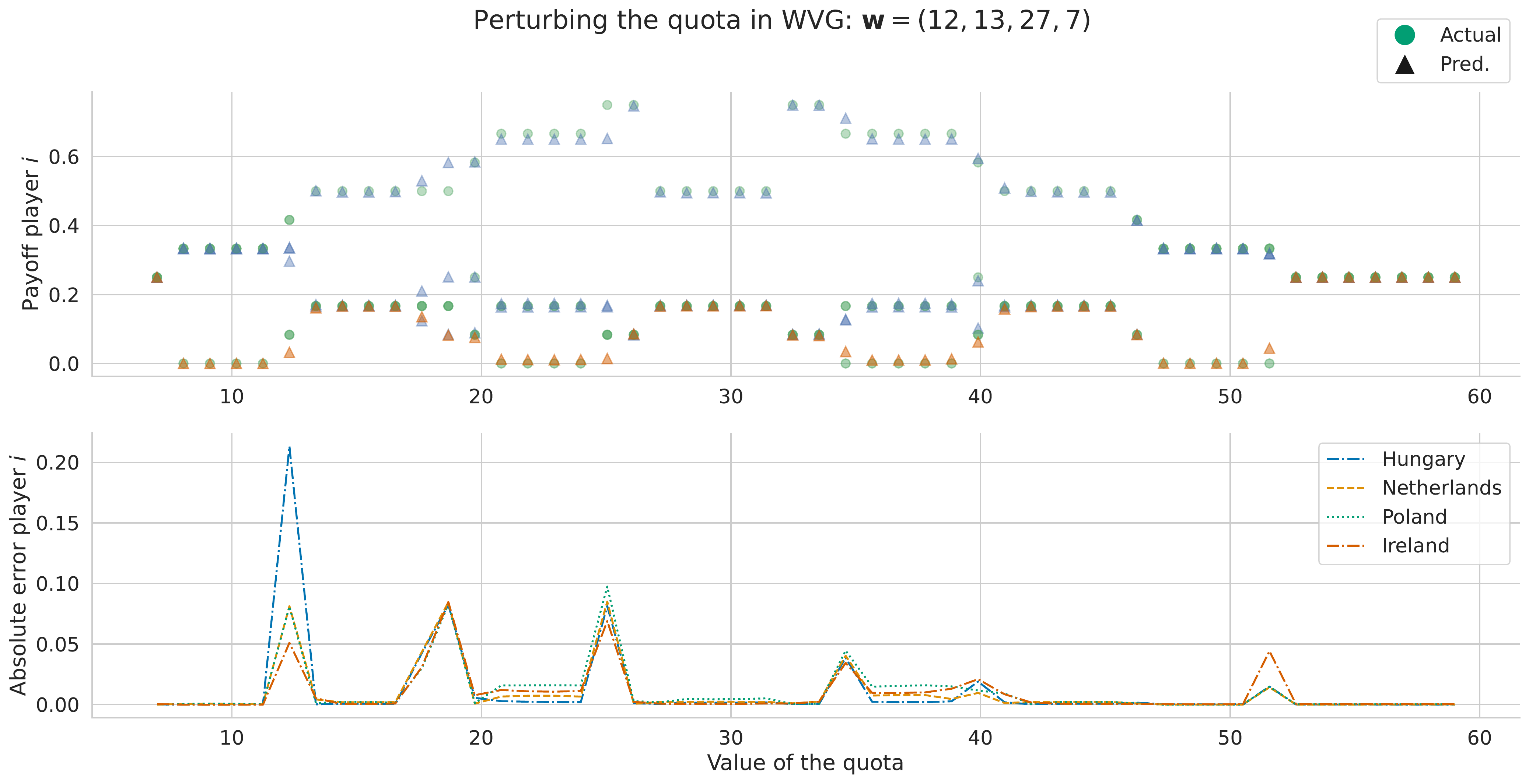}
    \caption{\textbf{Perturbing the quota in a voting game.} \textit{Top}: Model predictions and actual payoffs for a four-player voting game with weight vector $\mathbf{w} = (12, 13, 27, 7)$. \textit{Bottom}: Absolute errors between the individual member state payoffs. Note that these are a large number of transition points in the solutions.}
    \label{fig:quota-perturbation}
\end{figure}

We perform a similar analysis, this time perturbing the quota. Figure \ref{fig:quota-perturbation} depicts the same game with the original weights fixed, increasing the value of the quota from small to the sum of the weights (notice that this is the same experiment we performed in Section \ref{sec:capture_disc}). The value of the quota determines the set of winning coalitions, thereby dictating the payoff of each player. As such, perturbing in this game yields a large number of discontinuous jumps in the member states solutions compared to our previous analysis perturbing the weight of Hungary. Our model is able to capture the overall changes in the solutions but has difficulty with capturing large jumps in payoffs for small quota increments.

\begin{figure}[h!t]
    \centering
    \includegraphics[width=1\textwidth]{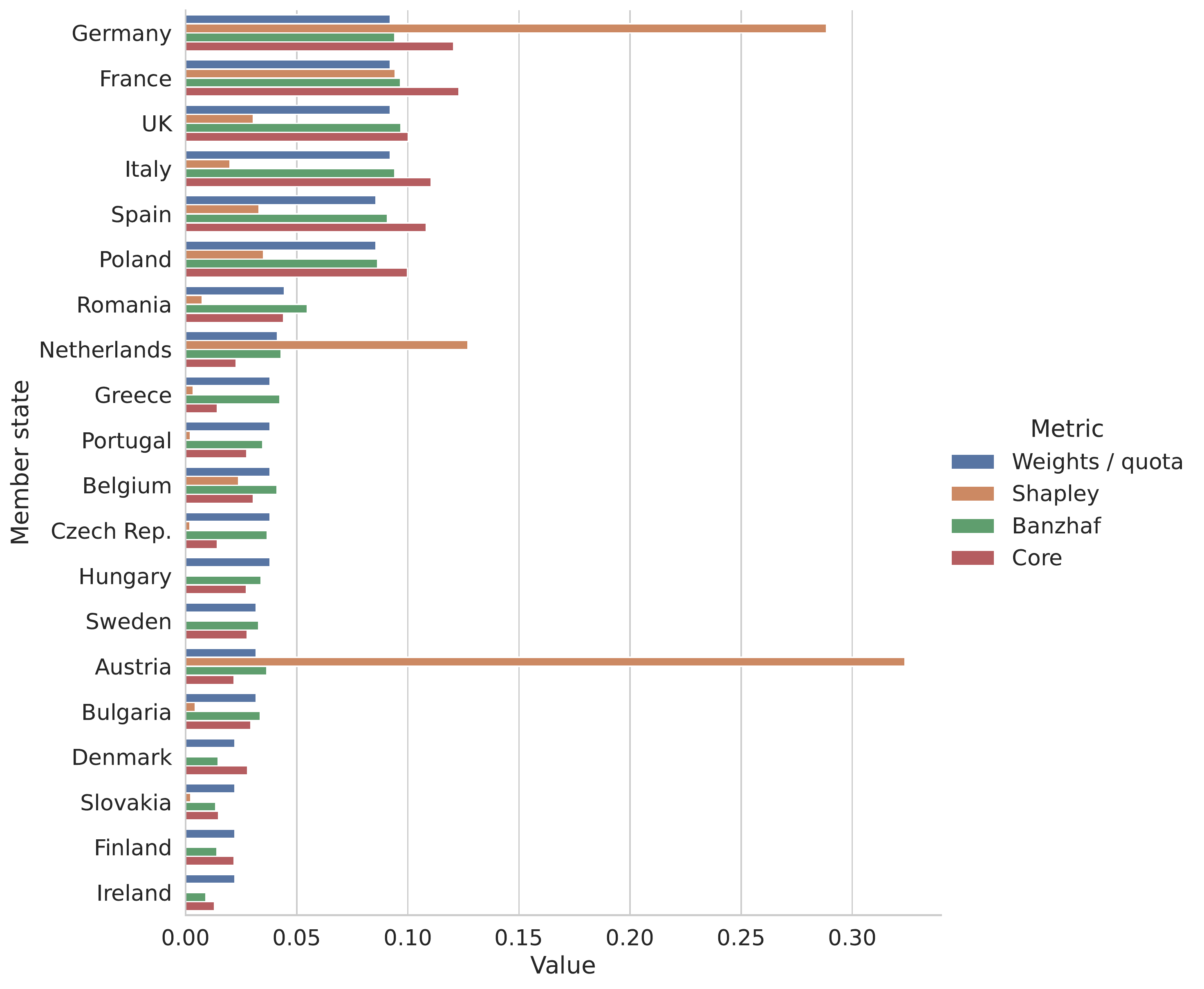}
    \caption{\textbf{Allocating payoffs to member states in the EU council with variable models}. A comparison of the allocated payoffs over solution concepts.}
    \label{fig:var-eu-application}
\end{figure}

Our variable models models have been trained on games up to ten players and can be applied to instantly predict the payoffs of member states in larger councils. To illustrate this, we take all member states in Table \ref{tab:EU-council-weights} and use our variable models to predict the payoffs for each, again setting the quota as the majority vote. We report the prediction quality of our models with respect to the ground truth (in Mean Mean Absolute Error), which is 0.013 for Shapley, 0.004 for Banzhaf, and 0.05 for the least core. Figure \ref{fig:var-eu-application} compares the solutions across member states and solution concepts.

While there is room for improvement in terms of prediction quality, our approach provides a useful tool for obtaining quick estimates of the voting power in medium size games ($10 \leq n < 50$). Furthermore, while approximation methods exist for Shapley and Banzhaf, there are currently no effective methods for estimating payoffs in the least core. This solution concept is more challenging as the solutions are obtained by solving a Linear Program with a hard constraints. Thus, solving a game with 15 or even 20 players contains on the order of $2^{15}$ or $2^{20}$ constraints, which slow down the solvers dramatically. Here, we can apply the same technique, solve for the small/medium size-games, then leverage our neural network approach to tackle a higher numbers of players. In other words, provided a solver that can handle up to $M$ players, we can train a neural network on this number of players, allowing us to extend the number of players even beyond $M$, possibly up to $2M$.

\clearpage
\section{Real World Applications of Weighted Voting Games}

Weighted voting games, and cooperative game theory in general, provide valuable tools for reasoning about the allocation of collective outcomes to individual agents. As such, their applicability span numerous domains. In this section, we highlight several real-world applications of our work. In all examples, our framework can be applied to both speed up and enhance the scalability. \citet{dinar1992evaluating} utilize solution concepts from cooperative game theory -- including the Core, the Shapley Value, the
Nucleolus and Nash -- to allocate water resources to competing parties in the presence of scarcity. More recently, \citet{mirzaei2022evaluation} also performed an extensive evaluation of both cooperative and non-cooperative game theoretic approach for water
management in trans-boundary river basins.

Moreover, cooperative game theory has proven useful in analyzing the costs of joint activities \cite{young1994cost}. \citet{vazquez1997owen} combine a model of airport games with the Shapley value to fairly distribute the aircraft landing fees among airlines. When an airport is build, the size of the runway is chosen to match the largest designed airplaine. The authors address how to fairly divide costs between airlines provided the different designs and sizes of their planes.
 
\citet{bistaffa2017cooperative} applied cooperative game theory to analyze the social ridesharing problem. Mobility platforms such as Lyft allow their users to share their trip with nearby users in real-time, providing a more environmentally friendly and cost effective alternative to commuting alone. They study how the costs of the ride should be divided among passengers. Another application is in supply chain management, where \citet{fiestras2011cooperative} study how to coordinate actions between suppliers, manufactorers, retailers and customers.

\end{document}